\documentclass{article} 
\usepackage{iclr2027_conference,times}

\usepackage{amsmath,amsfonts,bm}

\def\eqref#1{equation~\ref{#1}}

\def\1{\bm{1}}

\DeclareMathAlphabet{\mathsfit}{\encodingdefault}{\sfdefault}{m}{sl}
\SetMathAlphabet{\mathsfit}{bold}{\encodingdefault}{\sfdefault}{bx}{n}

\usepackage{hyperref}
\usepackage{url}

\usepackage{graphicx}
\usepackage{bm}
\renewcommand{\vec}[1]{\bm{#1}}

\title{Spatial But Not Temporal Reward Sparsity Drives Slow Policy Learning}

\author{%
  John J. Vastola, Kanaka Rajan \\
Department of Neurobiology, Harvard Medical School \\
Kempner Institute for the Study of Natural and Artificial Intelligence, Harvard University \\
  \texttt{\{john\_vastola, kanaka\_rajan\}@hms.harvard.edu} \\
}

\iclrfinalcopy 
\begin{document}

\maketitle
\pagestyle{plain}

\begin{abstract}
A primary difference between reinforcement and supervised learning is that rewards are generally sparse in both space and time: few actions yield high reward (spatial sparsity), and rewards may only be provided after temporally extended sequences of actions (temporal sparsity). How these factors impact policy learning dynamics is undertheorized, especially because it is difficult to design a theoretical setting interesting enough to capture relevant phenomenology, but simple enough to treat analytically. In this paper, we propose and analytically solve a minimal model of learning to pursue a moving target which naturally incorporates both spatial and temporal reward sparsity. We show mathematically that spatial sparsity can produce exponential slowdowns in learning, and that temporal sparsity is comparatively benign, only yielding a comparatively short additional phase of pursuit trajectory refinement. We also show that the presence of highly variable environmental state features can---without complex function approximators---produce an implicit curriculum effect which makes learning from sparse rewards polynomially rather than exponentially slow. We provide a wealth of relevant calculations, whose utility is in part that they sharpen our intuitions about reward sparsity.
\end{abstract}

\section{Introduction}

Reward sparsity is one of the primary factors that makes reinforcement learning (RL) \citep{sutton2018reinforcement} generally more difficult than supervised learning \citep{andrychowicz_hindsight_2017,hare_2019,ladosz2022exploration}. All else being equal, agents that are provided more frequent and informative feedback about which actions are `good' ought to learn more quickly. But there are multiple types of reward sparsity, and how these affect learning is poorly understood. 

The first, which we call \textit{spatial reward sparsity}, concerns the specificity of `good' actions: few actions yield high reward. This is true, for example, in robotics and continuous control settings with large action spaces. The second, which we call \textit{temporal reward sparsity}, concerns how often agents are rewarded in temporally extended tasks: reward may only be received after sequences of actions, which produces a temporal credit assignment problem. Episodic tasks where agents receive a reward only at the end, and not at any intermediate time step, are sparse in this sense. This is true of most games, since without auxiliary rewards, agents only receive a reward signal (a win or loss) at the end. Importantly, either kind of reward sparsity can be present without the other.

How do these two types of reward sparsity affect policy learning? Does one cause more difficulties, in the sense of making learning harder or slower, than the other? Existing theoretical work \citep{saglietti_analytical_2022,lee2024animals,patel2025rl} does not usually distinguish between them, and so is inconclusive---in part because it is generally difficult to conduct an exhaustive mathematical analysis unless the setting is taken to be extremely simple.

In this paper, we study the impact of spatial and temporal reward sparsity in the context of a tractable mathematical model of learning to pursue a moving target. This setting is natural because it is easy to implement both types of reward sparsity: the agent must overlap with the target to receive high reward (\textit{spatial sparsity}), and is rewarded mostly based on their final position rather than individual actions (\textit{temporal sparsity}). Although the model we consider is simple---it is linear in various senses---its phenomenology is surprisingly rich, and one can precisely characterize various aspects of its learning dynamics. Overall, we find that spatial reward sparsity determines the main bottleneck of learning when both spatial and temporal sparsity are present. Temporal sparsity mainly yields a separation between when the agent learns to reliably reach the target, and when they refine their trajectory en route to the target, but the convergence rate of the refinement phase remains exponentially fast regardless of the number of time steps. Meanwhile, unless something explicitly or implicitly addresses it, spatial sparsity can yield exponential slowdowns in learning speed. Aside from introducing a theoretical setting in which spatial and temporal reward sparsity can be disentangled and analytically studied, our main contributions are as follows:
\begin{itemize}
\item \textbf{Spatial sparsity yields slow learning.} We show sparse rewards can cause small differences in initial conditions to produce (exponentially) large increases in learning time, and that this issue worsens with increasing spatial, but not temporal, task dimensionality. (Sec. \ref{sec:spatial})
\item \textbf{Temporal sparsity yields trajectory refinement.} We show that temporal sparsity causes agents to learn to reach the target before they optimize how they get to it, and we connect the delay between these phases of learning to the eigenvalues of a certain matrix. (Sec. \ref{sec:temporal})
\item \textbf{Learning dynamics yields implicit curriculum.} We show that when environmental state variance is high, learning involves transient weight increases that effectively increase exploration and make learning polynomially rather than exponentially slow. (Sec. \ref{sec:curr} and \ref{sec:implicit})
\end{itemize}
Together, these results quantitatively sharpen our intuitions about how sparsity slows learning. They also show that slowdowns can be overcome by explicit (Sec. \ref{sec:curr}) or implicit (Sec. \ref{sec:implicit}) mechanisms.

\section{Mathematical formulation}
\label{sec:form}

In this paper, we study reward sparsity in the context of an episodic target pursuit task inspired by \cite{vastola2026solvable} and \cite{vastola2026effectiveness}: the agent must pursue a possibly moving target in $D \geq 1$ spatial dimensions. Our central concern in designing this theoretical setting was incorporating both spatial and temporal reward sparsity while keeping the dynamics of policy learning analytically tractable. Below, we introduce and motivate the task.

\paragraph{Task structure.} Each episode lasts for a fixed amount of time $T > 0$ and consists of $M$ time steps of length $\Delta t := T/M$. First, the agent observes a \textit{state vector} $\vec{z} := (z_1, ..., z_F)^T \in \mathbb{R}^F$ (where $F \geq 1$ is the number of features) that provides information about the target. While we treat this feature vector as an abstract quantity, it is intended to model, e.g., the target's velocity vector. Next, the agent submits a temporal sequence of $M$ velocities $\vec{u}_0, \vec{u}_1, ..., \vec{u}_{M-1} \in \mathbb{R}^D$, after which they arrive at an \textit{endpoint} $\vec{x}_T := \sum_{t = 0}^{M-1} \vec{u}_t \Delta t \in \mathbb{R}^D$ that corresponds to their integrated velocity vector. Note that the task is \textit{open-loop} since actions do not affect target behavior, which only depends on $\vec{z}$. 

Finally, they are rewarded according to a return $G(\vec{u}_{0:T}, \vec{z})$ which depends on both state and the sequence $\vec{u}_{0:T}$ of control velocities. For simplicity, we assume that the state vector is sampled independently from $\vec{z} \sim \mathcal{N}(\vec{0}, \vec{\Sigma})$ during each episode, where $\vec{\Sigma}$ is a fixed $F \times F$ covariance.

\paragraph{Reward structure.} The agent's goal is for its endpoint $\vec{x}_T := \sum_{t = 0}^{M-1} \vec{u}_t \Delta t$ to match the \textit{target endpoint} $\vec{f}_T(\vec{z}) \in \mathbb{R}^D$. We assume the return they receive has two components: an \textit{effort term} which penalizes high velocities $\vec{u}_t$, and a \textit{target proximity term} that incentivizes reaching the target:
\begin{equation} \label{eq:return}
G( \vec{u}_{0:T}, \vec{z} ) = G_{\text{effort}} + G_{\text{target}} =  \sum_{t=0}^{M-1} -\frac{\alpha}{2} \Vert \vec{u}_t \Vert_2^2   \ \Delta t +  g \left( \frac{\kappa}{2 \pi} \right)^{D/2} e^{- \frac{\kappa}{2} \Vert \vec{x}_T - \vec{f}_T(\vec{z}) \Vert_2^2}  \ .
\end{equation}
Here, $\alpha > 0$ controls the cost of each velocity input $\vec{u}_t$, $g > 0$ controls the overall size of the target proximity term, and $\kappa$ controls the \textit{sparsity} of rewards. For large $\kappa$, the second term essentially becomes a delta function centered on the target; for small $\kappa$, it becomes effectively quadratic. A major objective of our work is to characterize the impact of large $\kappa$ on learning dynamics. Note that the return from \cite{vastola2026effectiveness} can be viewed as a special case of ours with only one time step.


This task exhibits \textbf{spatial sparsity} in the sense that the target is localized in space, and the agent must end the episode within a distance of $1/\sqrt{\kappa}$ from it to receive high reward. It exhibits \textbf{temporal sparsity} in the sense that there is a credit assignment problem: most reward depends not on single actions, but essentially on the \textit{average} action, since this is what determines $\vec{x}_T := \sum_{t = 0}^{M-1} \vec{u}_t \Delta t$.

\paragraph{Target and agent movement structure.} We assume that the target endpoint depends linearly on $\vec{z}$, i.e., $\vec{f}_T(\vec{z}) = \vec{b}^f + \vec{W}^f \vec{z}$, where $\vec{b}^f \in \mathbb{R}^D$ is the \textit{target bias} and $\vec{W}^f \in \mathbb{R}^{D \times F}$ is the \textit{target weight matrix}. The agent's average action is assumed to have a similar form, with $\bar{\vec{u}}_t(\vec{z}) = \frac{1}{T} \vec{b}_t + \frac{1}{T} \vec{W}_t \vec{z}$; the control velocity is determined by a \textit{bias} $\vec{b}_t \in \mathbb{R}^D$ and \textit{weight matrix} $\vec{W}_t \in \mathbb{R}^{D \times F}$, both of which are learned. Note that \textit{actions associated with different time steps depend on different learned parameters} (for example, $\vec{b}_t$ vs $\vec{b}_{t'}$), which is unrealistic but important for tractability.

We assume that the agent's policy $\pi(\vec{u}_{0:T} | \vec{z})$ also involves \textit{exploration noise}: during each time step, the action $\vec{u}_t$ is sampled independently of other actions according to a normal distribution:
\begin{equation} \label{eq:policy_def}
\pi(\vec{u}_{0:T} | \vec{z}) = \prod_{t = 0}^{M-1} \pi( \vec{u}_t | \vec{z}) \hspace{1in} \pi(\vec{u}_t | \vec{z}) = \mathcal{N}(\vec{u}_t ; \bar{\vec{u}}_t(\vec{z}), \frac{1}{\Delta t} \vec{B}_t^{-1}) \ .
\end{equation}
The positive semidefinite matrices $\vec{B}_t^{-1} \in \mathbb{R}^{D \times D}$ control how exploratory the agent is, and are assumed \textit{fixed} rather than \textit{learned}; in Sec. \ref{sec:curr}, we will show that choosing them carefully can speed up learning.

\paragraph{Policy learning dynamics.} We study an agent which learns a policy for performing this pursuit task via episodic REINFORCE \citep{williams_simple_1992,sutton_policy_1999}. If we assume a small learning rate and a large number of episodes, the learning dynamics associated with a parameter $\theta_t$ of the time step $t$ mean action $\bar{\vec{u}}_t(\vec{z})$ can be approximately described (see Appendix \ref{app:dyn_deriv}) by the ordinary differential equation (ODE)
\begin{equation}
\frac{d}{d\tau} \theta_t = \frac{\eta T}{\Delta t}   \ \mathbb{E}_{\vec{z}}\left\{ \frac{\partial \bar{\vec{u}}_t(\vec{z})^T}{\partial \theta_t} \vec{B}_t \Delta t \ \mathbb{E}_{\vec{u}_{0:T} | \vec{z}}\left[ \ (\vec{u}_t - \bar{\vec{u}}_t(\vec{z})) \ G(\vec{u}_{0:T}, \vec{z}) \right]  \right\} \ ,
\end{equation}
where $\eta > 0$ is the (effective) learning rate. Although we neglect value learning for simplicity, it would play no role in this setting, since (at least when value is used as a baseline, as in actor-critic methods) value functions do not influence \textit{average} updates \citep{sutton2018reinforcement_chapter}. Value functions are usually used to reduce variance in learning dynamics, but---under the assumptions that permit the above ODE description---there is no variance in parameter updates.

\section{Spatial reward sparsity produces learning slowdown}
\label{sec:spatial}

We begin by studying \textit{pure bias learning}, in which there are no features (i.e., $F = 0$), so $\bar{\vec{u}}_t = \vec{b}_t/T$ and $\vec{f}_T = \vec{b}^f$. This case is simple while illustrating the basic difficulties associated with sparse rewards---and as we discuss in Sec. \ref{sec:implicit}, it is actually much \textit{harder} to learn than the case with features. Our first question is this: how much do sparser rewards (i.e., higher $\kappa$) slow down learning?

\paragraph{Endpoint learning dynamics.} Let $\vec{b} := \sum_{t = 0}^{M-1} \vec{b}_t \frac{\Delta t}{T}$ denote the average learned \textit{endpoint}, which ought to be close to the target endpoint $\vec{b}^f$ by the end of learning. After performing the average over the agent's action sequence $\vec{u}_{0:T}$ (see Appendix \ref{app:dyn_deriv}), we find that learning dynamics depend on 
\begin{equation} \label{eq:phit_bias}
p_{\text{hit}}( \vec{b} ) := \sqrt{\frac{\det\tilde{\vec{B}}}{(2 \pi)^D}} e^{- \frac{1}{2} (\vec{b} - \vec{b}^f)^T \tilde{\vec{B}} (\vec{b} - \vec{b}^f)} \hspace{0.1in} \text{ where } \hspace{0.1in} \tilde{\vec{B}}^{-1} :=  \frac{1}{\kappa} \vec{I} + \sum_{t=0}^{M-1} \vec{B}^{-1}_t \Delta t  \in \mathbb{R}^{D \times D} \ ,
\end{equation}
a quantity that we call the \textit{hitting probability}. This quantity modulates the impact of reward sparsity: when it is small, the agent receives little informative reward signal, and learning takes significantly longer. The matrix $\tilde{\vec{B}}^{-1}$ determines the \textit{effective target size} along each direction of $\mathbb{R}^D$: if it is smaller along a given axis, the agent must stop closer to the target to receive appreciable reward. 


We derive the learning dynamics of $\vec{b}_t$ in Appendix \ref{app:dyn_deriv}. But because the sparse reward term depends \textit{only} on the learned endpoint $\vec{b} := \sum_{t = 0}^{M-1} \vec{b}_t \frac{\Delta t}{T}$, and not on any specific $\vec{b}_t$, these dynamics are extremely similar to those of the endpoint. We will find it convenient to track $\vec{b}$ dynamics directly:
\begin{equation} \label{eq:end_dyn}
\frac{d}{d\tau} \vec{b} = \eta \left\{  - \frac{\alpha}{T} \vec{b}   + g \tilde{\vec{B}} ( \vec{b}^f - \vec{b}) \ \sqrt{\frac{\det\tilde{\vec{B}}}{(2 \pi)^D}} e^{- \frac{1}{2} (\vec{b} - \vec{b}^f)^T \tilde{\vec{B}} (\vec{b} - \vec{b}^f)}  \right\}  \ .
\end{equation}
The most salient characteristic of this $D$-dimensional dynamical system is that it has \textit{two regimes}, which are determined by whether the  bias error $E := (\vec{b} - \vec{b}^f)^T \tilde{\vec{B}} (\vec{b} - \vec{b}^f)$ is large or small. 





\paragraph{Dense reward regime.} When $E$ is somewhat less than $1$, so that $\vec{b}$ is close to the target endpoint, 
\begin{equation} \label{eq:spatial_dense}
\frac{d}{d\tau} \vec{b} \approx \eta \left[  - \frac{\alpha}{T} \vec{b}   + g \tilde{\vec{B}} ( \vec{b}^f - \vec{b}) \ \sqrt{\det\tilde{\vec{B}}/(2 \pi)^D} \right] \ ,
\end{equation}
which implies that $\vec{b}$ converges exponentially quickly to the (unique) optimum, which is near $\vec{b}^f$ if $\alpha/T$ is small. This occurs on a time scale determined by the eigenvalues of $\frac{\alpha}{T} \vec{I} + g \sqrt{\det\tilde{\vec{B}}/(2 \pi)^D} \tilde{\vec{B}}$. Exponentially fast convergence means that learning in this regime is `easy'.

\begin{figure} 
  \centering
  \includegraphics[width=\textwidth]{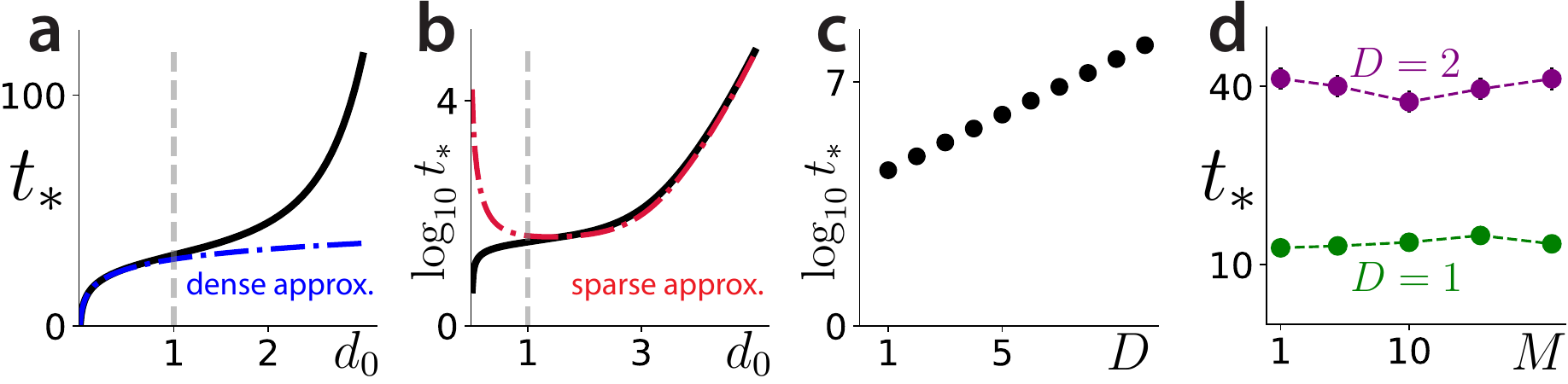}
  \caption{\textbf{Spatial sparsity produces exponential slowdown.} Bias units are chosen so that $\tilde{\beta} = 1$. \textbf{a.} The convergence time $t_*$ associated with endpoint learning dynamics (black; see Eq. \ref{eq:conv_gen}) scales poorly with the initial bias error $d_0$. For small enough $d_0$, it is approximately logarithmic in $d_0$ (blue curve); $d_0 = 1$ (gray line) is where the approximation breaks down. \textbf{b.} The convergence time is well-approximated as exponential (red; see Eq. \ref{eq:conv_slow}) when $d_0$ is large. \textbf{c.} Log-convergence-time scales linearly with the spatial dimensionality $D$ under typical assumptions. \textbf{d.} Convergence time is insensitive to the number of time steps $M$ under typical assumptions. This panel shows an empirical result ($N = 500$ seeds) for $D = 1$ (green) and $D = 2$ (purple); standard errors are shown but small.}
  \label{fig:space_sparse}
\end{figure}

\paragraph{Sparse reward regime.} When $E$ is somewhat \textit{greater} than $1$, so that $\vec{b}$ is far from the target endpoint, the sparse reward term is exponentially suppressed and learning greatly slows down. If we make the simplifying assumption that $\tilde{\vec{B}} = \tilde{\beta} \vec{I}$ is isotropic, and assume $\alpha/T$ is small enough to be negligible, we can find (see Appendix \ref{app:conv}) that the bias error $E$ evolves as
\begin{equation}
\frac{d}{d\tau} E = - 2 \eta g \tilde{\beta} \left(  \frac{\tilde{\beta}}{2 \pi} \right)^{D/2} \ E e^{- E/2} \ .
\end{equation}
This ODE implies a convergence time result: the time to go from an initial error $\tilde{\beta} \Vert \vec{b}^0 - \vec{b}^f \Vert_2^2$ to a final (small) error $\epsilon^2$ equals
\begin{equation} \label{eq:conv_gen}
t_* =  \frac{1}{2 \eta g \tilde{\beta}} \left( \frac{2 \pi}{\tilde{\beta}} \right)^{D/2} \left[  \ \text{Ei}\left( \frac{\tilde{\beta}}{2} \Vert \vec{b}^0 - \vec{b}^f \Vert_2^2  \right) - \text{Ei}(  \frac{1}{2} \epsilon^2   ) \ \right] \ ,
\end{equation}
where $\text{Ei}(x) := \int_{-\infty}^x \frac{e^y}{y} dy$ is the \textit{exponential integral function}. For small initial errors, $t_*$ depends logarithmically on the initial error (Fig. \ref{fig:space_sparse}a). But since $\text{Ei}(x) \approx e^x/x$ for large arguments, when the initial endpoint is somewhat far from the target $\vec{b}^f$, 
\begin{equation} \label{eq:conv_slow}
t_* \approx  \frac{1}{\eta g \tilde{\beta}^2} \left( \frac{2 \pi}{\tilde{\beta}} \right)^{D/2}   \ \frac{e^{ \frac{\tilde{\beta}}{2} \Vert \vec{b}^0 - \vec{b}^f \Vert_2^2}}{\Vert \vec{b}^0 - \vec{b}^f \Vert_2^2} \ ,
\end{equation}
where we have neglected the (much smaller) $\epsilon$-dependent term. This result is significant because it says the convergence time scales \textit{exponentially} with the initial bias error: small increases in the initial error will produce large increases in learning time (Fig. \ref{fig:space_sparse}b). We conclude that reward sparsity---$\kappa$ being large, which makes $\tilde{\beta}$ large---\textit{substantially slows learning}.

\paragraph{Spatial sparsity, not temporal sparsity, controls exponential slowdown.} From Eq. \ref{eq:conv_slow}, it is clear that the initial endpoint error largely controls the slowness of learning. How does this depend on the spatial dimensionality $D$ and number of time steps $M$?

To ensure that the endpoint $\vec{x}_T := \sum_{t = 0}^{M-1} \vec{u}_t \Delta t$ is well-behaved for large $M$, it is typical to initialize policy parameters so that the variance of $\vec{x}_T$ is independent of $M$. For example, choosing $b^0_{t i} \sim \mathcal{N}( 0, \xi^2 (T/\Delta t) )$ for all $t = 0, 1, ..., M-1$ and $i = 1, ..., D$, we have
\begin{equation}
\mathbb{E}_{\vec{b}^0_{0:T}}[ \vec{x}_T \vec{x}_T^T ] = \sum_{t, t'} \frac{(\Delta t)^2}{T^2} \mathbb{E}[ \vec{b}^0_t (\vec{b}^0_{t'})^T ] = \sum_{t, t'} \frac{(\Delta t)^2}{T^2} \delta_{t, t'} \xi^2 \frac{T}{\Delta t} \vec{I} = \xi^2 \vec{I} \ .
\end{equation}
If we assume that components of the target location are chosen independently according to $b^f_i \sim \mathcal{N}(0, \xi_f^2)$, then a typical value of the initial error is
\begin{equation}
\tilde{\beta} \ \mathbb{E}[ \Vert \vec{b}^0 - \vec{b}^f \Vert_2^2 ] = \tilde{\beta} \left\{ \mathbb{E}[ \Vert \vec{b}^0 \Vert_2^2 ] + \mathbb{E}[\Vert \vec{b}^f \Vert_2^2]     \right\} = \tilde{\beta} D \left\{ \xi^2 + \xi_f^2    \right\} \ .
\end{equation}
If we heuristically write the initial error as $\tilde{\beta} D d_0^2$, we finally obtain a convergence time
\begin{equation} \label{eq:conv_base}
t_* \approx  \frac{1}{\eta g \tilde{\beta}^2 d_0^2} \left( \frac{2 \pi}{\tilde{\beta}} e^{ \tilde{\beta}  d_0^2} \right)^{D/2}   \ \frac{1}{D}
\end{equation}
which scales \textit{exponentially} in $D$ (Fig. \ref{fig:space_sparse}c), but does \textit{not} scale with $M$ (Fig. \ref{fig:space_sparse}d). We conclude that \textbf{spatial reward sparsity, not temporal reward sparsity, drives slow policy learning} in this model.

\section{Temporal reward sparsity produces refinement phase}
\label{sec:temporal}


It may seem strange that the exponential slowdown we just described does not depend at all on our task's temporal extent---intuitively, one expects tasks with more time steps to be harder to learn. In this section, we show that this is indeed the case, but that the difficulties produced by temporal reward sparsity are significantly less severe than those described in the previous section. 

\begin{figure} 
  \centering
 \includegraphics[width=\textwidth]{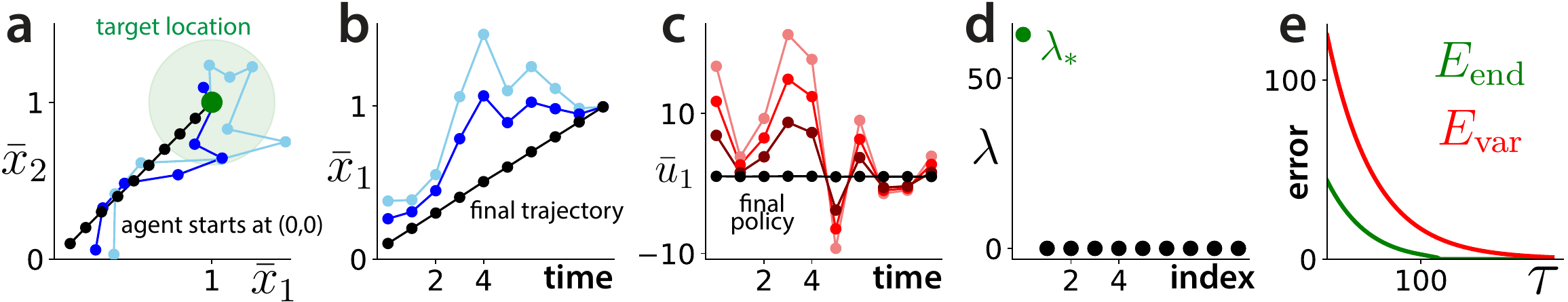}
  \caption{\textbf{Temporal reward sparsity delays trajectory refinement.} All simulations use $M = 10$ time steps and $D = 2$. \textbf{a.} Average learned trajectory (obtained by summing agent's selected velocities). Examples from three different parts of learning are shown: around when the agent first learns to hit the target (light blue), their trajectory is not straight; later, it becomes straighter (blue), and by the end of learning it is straight (black) since this minimizes the effort penalty. \textbf{b.} Same as in a., but trajectory (along the first direction only) visualized as a function of time step. \textbf{c.} As learning proceeds, the velocities the agent selects at each time step get increasingly similar (light red to dark red), and by the end they are identical (black). \textbf{d.} The matrix $\vec{A}$ controlling trajectory refinement has a `gapped' spectrum with one outlier eigenvalue. \textbf{e.} Endpoint error (green) and trajectory variance (red) converge at different rates; the agent learns to reach the target before optimizing their path.}
  \label{fig:temp_sparse}
\end{figure}

\paragraph{Endpoint is learned first.} Because the sparse reward term is determined only by the average endpoint $\vec{b} := \sum_{t = 0}^{M-1} \vec{b}_t \frac{\Delta t}{T}$, in the initial (sparse regime) phase of learning the individual $\vec{b}_t$ are relatively unconstrained. When $\vec{b}$ is sufficiently close to the target endpoint $\vec{b}^f$, so that we obtain \textit{dense regime} dynamics, the $\vec{b}_t$ evolve according to 
\begin{equation} \label{eq:traj_linear}
\frac{d}{d\tau} \vec{b}_t \approx \eta \left\{  - \frac{\alpha}{T} \vec{b}_t   + g \tilde{\vec{B}} ( \vec{b}^f - \vec{b}) \ \sqrt{\frac{\det\tilde{\vec{B}}}{(2 \pi)^D}} \right\} \ ,
\end{equation}
which is almost identical to Eq. \ref{eq:spatial_dense}. It is only in this second phase, after the endpoint $\vec{b}$ has been roughly learned (Fig. \ref{fig:temp_sparse}a-b), that individual $\vec{b}_t$ (i.e., the policy at specific times; Fig. \ref{fig:temp_sparse}c) are refined.

\paragraph{Matrix with a spiked spectrum controls rate of trajectory refinement.} Consider an eigendecomposition $\tilde{\vec{B}} = \vec{O} \vec{\Lambda}_B \vec{O}^T$ of $\tilde{\vec{B}}$, where $\vec{O}$ is orthogonal (i.e., $\vec{O} \vec{O}^T = \vec{I}$) and $\vec{\Lambda}_B$ is diagonal. Let $\tilde{\beta}_i := \Lambda_{B, ii}$ denote its $i$-th diagonal entry. Eq. \ref{eq:traj_linear} can be simplified by defining rotated variables $\tilde{\vec{b}}_t := \vec{O}^T \vec{b}_t$, since the $i$-th entry of this vector evolves according to (see Appendix \ref{app:temp})
\begin{equation} \label{eq:tilde_sys}
\frac{d}{d\tau} \tilde{b}_t = \eta \left\{  - \frac{\alpha}{T}  \tilde{b}_t    + g_i \left( \tilde{b}^f -  \sum_t \tilde{b}_t  \frac{\Delta t}{T} \right) \right\} \hspace{0.5in} g_i :=  g \sqrt{\det\tilde{\vec{B}}/(2 \pi)^D } \tilde{\beta}_i 
\end{equation}
where we have suppressed the $i$ index of $\tilde{b}_t$ and $\tilde{b}^f$ to ease notation. This reduction is useful because different spatial components of $\tilde{\vec{b}}_t$ evolve independently; the above ODE describes the evolution of $M$ parameters, each associated with a different time point. Define a matrix $\vec{A} := \frac{\alpha}{T} \vec{I} + g_i \frac{\Delta t}{T} \vec{1} \vec{1}^T$, a vector $\vec{y} := ( \tilde{b}_0, \tilde{b}_1, ..., \tilde{b}_{M-1})^T \in \mathbb{R}^M$, and $\vec{y}^* := g_i \tilde{b}^f \vec{A}^{-1} \vec{1}$. We can rewrite Eq. \ref{eq:tilde_sys} as the ODE
\begin{equation}
\frac{d}{d\tau} \vec{y} =  \eta \vec{A} (\vec{y}^* - \vec{y}) \ ,
\end{equation}
which we can solve to find that $\vec{y}(\tau) = \vec{y}^* + e^{- \eta \vec{A} \tau} [ \vec{y}(0) - \vec{y}^* ]$. It is interesting that $\vec{A}$ controls $\tilde{\vec{b}}_t$ convergence, because it has a `spiked' spectrum: it has an `outlier' eigenvector $\vec{1}$ whose eigenvalue 
\begin{equation}
\lambda_* := \frac{\alpha}{T} + g_i
\end{equation}
is larger than the other eigenvalues, which equal $\lambda_{\text{bulk}} := \alpha/T$ (Fig. \ref{fig:temp_sparse}d). But $\vec{1}$ being an eigenvector is equivalent to saying that the endpoint $\sum_t \tilde{b}_t \frac{\Delta t}{T}$, which corresponds to this eigenvector up to scaling, is learned earliest. Hence, the gap between endpoint learning and trajectory refinement is controlled by this eigenvalue gap, and in particular the relative sizes of $g_i$ and $\alpha/T$.

\paragraph{Trajectory error is reduced after bias error.} Another way of quantifying the difference between when the agent learns to reach the target, and when they learn to reach the target \textit{efficiently} (i.e., in a way which minimizes the effort term), is to track the total trajectory error, which decomposes as
\begin{equation}
E_{\text{tot}} := \sum_{i = 1}^D \sum_{t = 0}^{M-1} \frac{\Delta t}{T} ( b_{t i} - b^f_i )^2 =  \sum_{i = 1}^D  ( b_{i} - b^f_i )^2 +  \sum_{i = 1}^D \sum_{t = 0}^{M-1} \frac{\Delta t}{T} ( b_{t i} - b_i )^2 = E_{\text{end}} + E_{\text{var}} \ ,
\end{equation}
i.e., as a sum of the \textit{endpoint error} and the \textit{variance of actions}. It is the second term which causes the total error to shrink more slowly than the endpoint error (Fig. \ref{fig:temp_sparse}e); it converges on a time scale determined by $\lambda_{\text{bulk}}$. We conclude that \textbf{temporal reward sparsity only somewhat slows learning, since parameters still converge exponentially quickly to their optimal values} (after hitting).

\section{Optimal curricula avoid slow learning by increasing hitting }
\label{sec:curr}

\begin{figure}
  \centering
  \includegraphics[width=\textwidth]{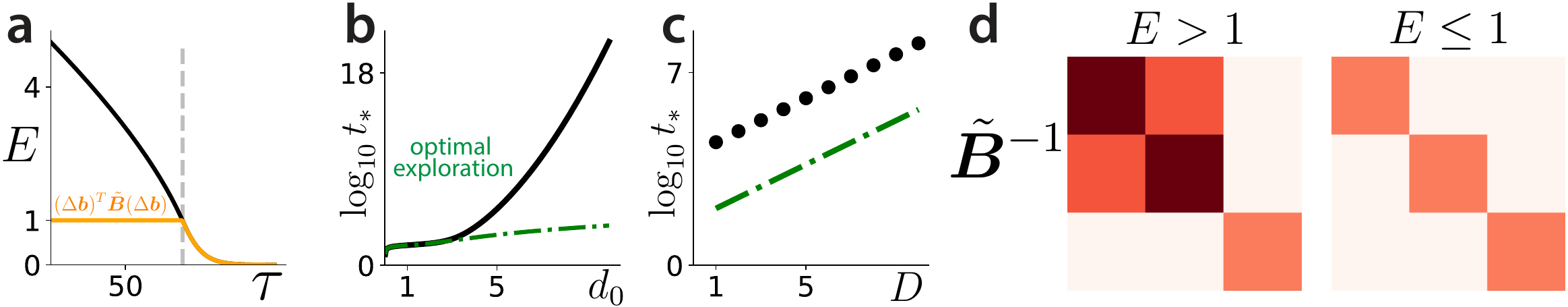}
  \caption{\textbf{Optimal curricula avoid slow learning by increasing hitting probability.} Bias units are chosen so that $\kappa = 1$. \textbf{a.} When the `optimal' exploration matrix $\vec{B}$ (see Eq. \ref{eq:B_opt}) is used, the endpoint error $E := \kappa \Vert \Delta \vec{b} \Vert_2^2$ decreases steadily, and the quantity $(\Delta \vec{b})^T \tilde{\vec{B}} (\Delta \vec{b})$ (orange line) is held fixed at $1$ until increasing exploration cannot usefully improve the hitting probability (gray line).   \textbf{b.} Using the optimal $\vec{B}$ improves the way the convergence time $t_*$ scales with the initial bias error $d_0$: it is cubic in $d_0$ (green; see Eq. \ref{eq:poly_slow}) rather than exponential (black; see Sec. \ref{sec:spatial}).  \textbf{c.} This result still scales exponentially with the spatial dimensionality $D$, but note the orders-of-magnitude improvement for any given $D$. \textbf{d.} While the variance-matching condition (Eq. \ref{eq:var_match}) is satisfied, the optimal $\tilde{\vec{B}}^{-1}$ is a scaled identity matrix plus a low-rank perturbation; when it cannot be satisfied, because $E \leq 1$, the optimal $\tilde{\vec{B}}^{-1}$ is a scaled identity matrix. Hotter colors indicate larger nonzero entries.}
  \label{fig:optimal}
\end{figure}


Is it possible to avoid the exponential slowdown due to (spatial) reward sparsity? The answer is \textit{yes}---by dynamically modulating task difficulty to the agent's current ability, it can be avoided. Here, we describe an approach which does this \textit{explicitly}, by changing how exploratory the agent is over time, which is a form of \textit{curriculum learning} \citep{bengio2009curriculum,wang2022survey}. Another approach, which we describe in Sec. \ref{sec:implicit}, accomplishes this \textit{implicitly} through learning dynamics. 

According to Eq. \ref{eq:end_dyn}, the matrix $\tilde{\vec{B}}^{-1} := [ (1/\kappa) \vec{I} + \sum_{t=0}^{M-1} \vec{B}^{-1}_t \Delta t ]$ determines the \textit{effective target size}, which in turn determines task difficulty, since learning is slow when the hitting probability is small. This means we can effectively modulate the hitting probability and the associated slowdown by choosing the exploration parameters $\vec{B}_t$ to \textit{ensure that the hitting probability remains large}. Define $\vec{B}^{-1} := \sum_{t=0}^{M-1} \vec{B}^{-1}_t \Delta t$, since this is the (controllable) parameter combination that determines $\tilde{\vec{B}}$. We are interested in adaptively choosing $\vec{B}^{-1}$ throughout learning in order to speed it up as much as possible. A key fact is that endpoint learning dynamics (Eq. \ref{eq:end_dyn}) perform gradient ascent on 
\begin{equation}
\frac{d}{d\tau} \vec{b} = \eta \frac{\partial G_{\text{eff}}(\vec{b}, \vec{B})}{\partial \vec{b}} \hspace{0.1in} \text{ where } \hspace{0.1in} G_{\text{eff}}(\vec{b}, \vec{B}) := - \frac{\alpha}{2 T} \Vert \vec{b} \Vert_2^2 + g \frac{e^{- \frac{1}{2} (\Delta \vec{b})^T [ (1/\kappa) \vec{I} + \vec{B}^{-1} ]^{-1} (\Delta \vec{b}) } }{\sqrt{(2 \pi)^D \det[ (1/\kappa) \vec{I} + \vec{B}^{-1}]}} \ ,
\end{equation}
which we call the \textit{effective return}. We can thus define an `optimal' $\vec{B}$ as one maximizing $G_{\text{eff}}$.\footnote{Note that this only works if we assume $\alpha/T$ is small. Otherwise, this optimization is not sensitive to the effort term, which can lead to the agent getting stuck in a spurious optimum.}



\paragraph{Optimal exploration maximizes hitting probability.} Maximizing $G_{\text{eff}}$ with respect to $\vec{B}$, subject to a positive semidefiniteness constraint, yields an `optimal' matrix (see Appendix \ref{app:optimal}; Fig. \ref{fig:optimal}d)
\begin{equation} \label{eq:B_opt}
\vec{B}^{-1}_*(\Delta \vec{b}) = \begin{cases}
\left( 1 - \frac{1}{\kappa \Vert \Delta \vec{b} \Vert_2^2}\right) (\Delta \vec{b}) (\Delta \vec{b})^T & \text{ if } \kappa \Vert \Delta \vec{b} \Vert_2^2 > 1 \\ 
\vec{0} & \text{ if } \kappa \Vert \Delta \vec{b} \Vert_2^2 \leq 1
\end{cases}  \ .
\end{equation}
This result is interesting, since it implies that (if $\kappa \Vert \Delta \vec{b} \Vert_2^2 > 1$) the \textit{variance-matching condition}
\begin{equation} \label{eq:var_match}
(\Delta \vec{b})^T \tilde{\vec{B}} (\Delta \vec{b}) = 1
\end{equation}
holds. This condition \textit{ensures the hitting probability remains large} (Fig. \ref{fig:optimal}a), as intuition suggests. Moreover, it has a reasonable interpretation: \textit{explore when it improves the hitting probability}.

\paragraph{Optimal exploration yields only polynomially slow learning.} The significance of Eq. \ref{eq:var_match} is that it produces (endpoint) learning dynamics (see Appendix \ref{app:optimal})
\begin{equation} \label{eq:bdyn_curr}
\frac{d}{d\tau} \vec{b} \approx- \frac{\eta g e^{-1/2}}{\kappa^{1/2}} \left( \frac{\kappa}{2 \pi} \right)^{D/2} \  \frac{\Delta \vec{b}}{\Vert \Delta \vec{b} \Vert_2^3}
\end{equation}
where we have assumed the effort term is negligibly small. The dynamics of the endpoint error $E :=  \kappa \Vert \Delta \vec{b} \Vert_2^2$ are then
\begin{equation}
\frac{d}{d\tau} E = -2 \kappa \eta g e^{-1/2} \left( \frac{\kappa}{2 \pi} \right)^{D/2} \  \frac{1}{\sqrt{E}}\ ,
\end{equation}
which implies a convergence time
\begin{equation} \label{eq:poly_slow}
t_* = \frac{1}{3}  \frac{e^{1/2}}{\kappa \eta g} \left( \frac{2 \pi}{\kappa} \right)^{D/2} \left[  \kappa^{3/2} \Vert \Delta \vec{b}^0 \Vert_2^3 - 1  \right] +  \frac{1}{2 \eta g \kappa} \left( \frac{2 \pi}{\kappa} \right)^{D/2} \left[  \text{Ei}(1/2) - \text{Ei}\left(  \frac{1}{2} \epsilon^2   \right)  \right] .
\end{equation}
Compare this result to our earlier (no curriculum) result, Eq. \ref{eq:conv_base}. Notably, the dependence on the initial error $\tilde{\beta} \Vert \Delta \vec{b}^0 \Vert_2^2$ is \textit{polynomial} rather than \textit{exponential} (Fig. \ref{fig:optimal}b-c). This implies, \textbf{given an appropriate exploration curriculum, sparse rewards need not produce exponential slowdowns.}

\section{Feature variability yields beneficial implicit curriculum}
\label{sec:implicit}
\begin{figure}
  \centering
  \includegraphics[width=0.6\textwidth]{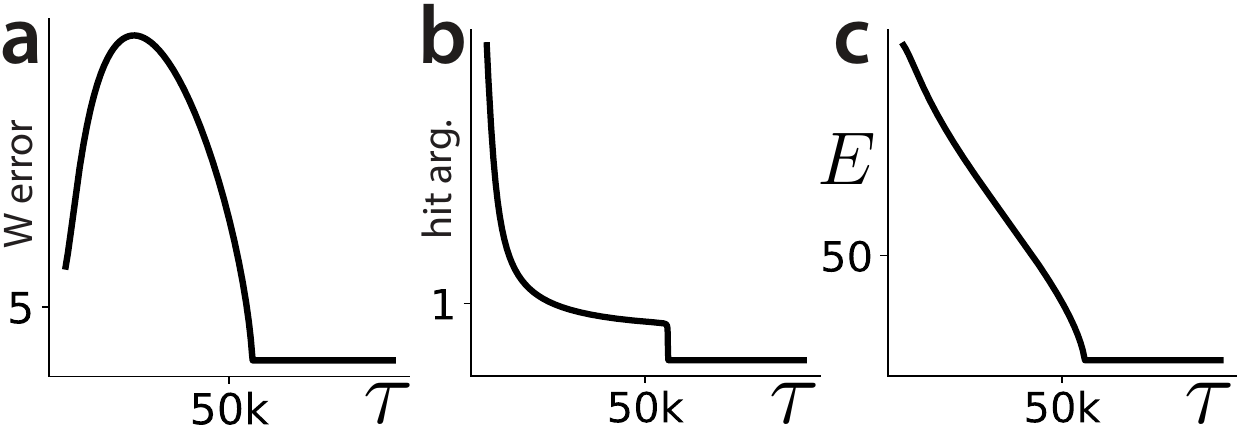}
  \caption{\textbf{Weight dynamics produces an implicit curriculum effect.} \textbf{a.} Example weight error trajectory. Note that it transiently increases, and then decreases. \textbf{b.} The transient increase in weights ensures that the hitting probability argument $(\Delta \vec{b})^T \vec{V}^{-1} (\Delta \vec{b})$ approaches $1$. \textbf{c.} Bias error decreases substantially around the time that the hitting probability argument is near $1$.}
  \label{fig:implicit}
\end{figure}

We are now ready to study the impact of reward sparsity on the more complex case, which involves $F \geq 1$ features and learning an (average) policy $\bar{\vec{u}}_t(\vec{z}) = \frac{1}{T} \vec{b}_t + \frac{1}{T} \vec{W}_t \vec{z}$. Because the task now involves episode-to-episode variability that comes from the state vector $\vec{z} \sim \mathcal{N}(\vec{0}, \vec{\Sigma})$, and $D \cdot F$ additional parameters to learn, one expects learning to be harder; however, we will show that learning is generally \textit{easier} than if there are no features. In particular, weight learning dynamics can produce an \textit{implicit curriculum learning effect} which avoids the exponential slowdown discussed in Sec. \ref{sec:spatial} and looks more like the explicit curriculum dynamics discussed in Sec. \ref{sec:curr} (see Fig. \ref{fig:implicit}). 

\paragraph{Learning dynamics.} In this more complex case, learning dynamics read (see Appendix \ref{app:dyn_deriv})
\begin{equation*}
\begin{split}
\frac{d}{d\tau} \vec{b}_t &= \eta \left\{  - \frac{\alpha}{T}  \vec{b}_t    + g \vec{V}^{-1} ( \vec{b}^f - \vec{b}) \frac{e^{- \frac{1}{2} (\Delta \vec{b})^T \vec{V}^{-1} (\Delta \vec{b})}  }{\sqrt{(2 \pi)^D \det \vec{V}}} \right\} \\
\frac{d}{d\tau} \vec{W}_t &= \eta \left\{   - \frac{\alpha}{T}  \vec{W}_t \vec{\Sigma}   + g \left[    \vec{I} - \vec{V}^{-1} (\Delta \vec{b}) (\Delta \vec{b})^T  \right] \vec{V}^{-1} (\vec{W}^f - \vec{W}) \vec{\Sigma}  \ \frac{e^{- \frac{1}{2} (\Delta \vec{b})^T \vec{V}^{-1} (\Delta \vec{b})} }{\sqrt{(2 \pi)^D \det \vec{V}}}    \right\}
\end{split}
\end{equation*}
where $\vec{W} := \sum_{t = 0}^{M-1} \vec{W}_t \frac{\Delta t}{T}$ is the weight matrix associated with the average learned endpoint, and
\begin{equation}
\vec{V} := \tilde{\vec{B}}^{-1}  +  (\vec{W} - \vec{W}^f) \vec{\Sigma} (\vec{W} - \vec{W}^f)^T \in \mathbb{R}^{D \times D}
\end{equation}
determines the \textit{effective target size}. Note that this expression is similar to what we had before (Eq. \ref{eq:phit_bias}), but involves a new weight-dependent term which \textit{grows} as the mismatch between $\vec{W}$ and $\vec{W}^f$ increases. In the remainder of this section, we show that this weight-dependent modulation of the effective target size produces an \textit{implicit curriculum learning effect} where agents transiently move weights \textit{away} from the optimal weights in order to increase their rate of hitting the target.

To ease analysis, we will make several simplifying assumptions: (i) the feature covariance matrix is isotropic, i.e., $\vec{\Sigma} = s \vec{I}$ for some $s > 0$; (ii) $s > 0$ is so large that the bias parameters are effectively frozen on the time scale of weight dynamics; and (iii) the effort coefficient $\alpha/T$ is small enough that it does not appreciably influence endpoint dynamics. Given these assumptions, the endpoint weight discrepancy $\Delta \vec{W} := \vec{W} - \vec{W}^f$ approximately evolves according to
\begin{equation} \label{eq:weight_simple}
\frac{d}{d\tau} \Delta \vec{W} \approx - \eta s g p_{\text{hit}}  \left[    \vec{I} - \vec{V}^{-1} (\Delta \vec{b}) (\Delta \vec{b})^T  \right] \vec{V}^{-1} (\Delta \vec{W})  \ .
\end{equation}

\paragraph{Weight dynamics have two fixed points, one of which implements implicit curriculum.} To understand weight dynamics, it is convenient to directly track the evolution of $\vec{V}$. Differentiating,
\begin{equation} \label{eq:V_dyn}
\frac{d}{d\tau} \vec{V} = - 2 \eta s g p_{\text{hit}} \left\{  \vec{I} - \frac{1}{\tilde{\beta}} \vec{V}^{-1} + \frac{1}{\tilde{\beta}}  \vec{V}^{-1} (\Delta \vec{b}) (\Delta \vec{b})^T \vec{V}^{-1}   - \frac{1}{2} \vec{V}^{-1} (\Delta \vec{b}) (\Delta \vec{b})^T -\frac{1}{2}  (\Delta \vec{b}) (\Delta \vec{b})^T \vec{V}^{-1}  \right\} .
\end{equation}
The matrix $\vec{V}$ stops changing when the right-hand side of this ODE equals zero, i.e., when
\begin{equation}
 \vec{I} - \frac{1}{\tilde{\beta}} \vec{V}^{-1} + \frac{1}{\tilde{\beta}}  \vec{V}^{-1} (\Delta \vec{b}) (\Delta \vec{b})^T \vec{V}^{-1}   = \frac{1}{2} \vec{V}^{-1} (\Delta \vec{b}) (\Delta \vec{b})^T +  \frac{1}{2} (\Delta \vec{b}) (\Delta \vec{b})^T \vec{V}^{-1} \ .
\end{equation}
This Riccati equation has two solutions. One corresponds to the obvious fixed point where $\vec{W} = \vec{W}^f$ (or in terms of $\vec{V}$, where $\vec{V}^{-1} = \tilde{\beta} \vec{I}$), whose presence is apparent even from Eq. \ref{eq:weight_simple}. But this fixed point becomes \textit{unstable} if $\tilde{\beta} \Vert \Delta \vec{b} \Vert_2^2 > 1$, and the alternative fixed point with
\begin{equation} \label{eq:vmatch_weight}
\vec{V}^{-1} = \tilde{\beta} \vec{I} +  \left[ \frac{1}{\Vert \Delta \vec{b}\Vert_2^2} - \tilde{\beta} \right] \frac{(\Delta \vec{b}) (\Delta \vec{b})^T}{\Vert \Delta \vec{b} \Vert_2^2}
\end{equation}
becomes stable (see Appendix \ref{app:implicit}). Importantly, at this fixed point we have $(\Delta \vec{b})^T \vec{V}^{-1} (\Delta \vec{b}) = 1$, which is \textit{precisely the same variance-matching condition we observed in Sec. \ref{sec:curr}}. Hence, at least at steady state, weight dynamics effectively increase the hitting probability in a manner akin to the explicit curriculum we studied in Sec. \ref{sec:curr}.

\paragraph{Error-aligned weight component gets magnified and other components get suppressed.} What happens away from these fixed points? To simplify our analysis, we will assume that $(\Delta \vec{b})$ is an eigenvector of $\vec{V}$; this is true at both fixed points, and if this is not true, one can show that $\vec{V}$ generally evolves so that this becomes true (see Appendix \ref{app:implicit}). Let $\lambda$ denote the eigenvalue associated with $\Delta \vec{b}$, and $\lambda_{\perp}$ an eigenvalue associated with a different eigenvector. Eq. \ref{eq:V_dyn} implies
\begin{equation}
\begin{split}
\frac{d}{d\tau} \lambda &= - 2 \eta s g p_{\text{hit}} \ \left(  1 - \frac{1}{\tilde{\beta}} \frac{1}{\lambda}  \right) \left(  1 - \frac{\Vert \Delta \vec{b} \Vert_2^2}{\lambda} \right)  \hspace{0.5in} \frac{d}{d\tau} \lambda_{\perp} = - 2  \eta s g p_{\text{hit}} \left( 1 - \frac{1}{\tilde{\beta} \lambda_{\perp}} \right) \ .
\end{split}
\end{equation}
The error-associated eigenvalue has two fixed points, one of which corresponds to the nontrivial one, whereas the other eigenvalues have a single fixed point at $\lambda_{\perp} = 1/\tilde{\beta}$. Hence, as long as $\tilde{\beta} \Vert \Delta \vec{b} \Vert_2^2 > 1$, one eigenvalue gets magnified while others get suppressed, with the net effect being that $(\Delta \vec{W}) (\Delta \vec{W})^T$ becomes rank-one and error-aligned. Overall, we recover a picture of learning which is similar to, but slightly more complicated than the one studied in Sec. \ref{sec:curr}: there is an initial phase where weight dynamics move so that the variance-matching condition holds (Eq. \ref{eq:vmatch_weight}), and then a phase where bias parameters effectively follow Eq. \ref{eq:bdyn_curr}. This continues until the hitting probability becomes small, which causes the cycle to repeat. Sparse rewards only produce a polynomial rather than exponential slowdown, with Eq. \ref{eq:poly_slow} holding in the infinite time scale separation limit. 

\section{Discussion}

In this work, we studied a model of learning to pursue a moving target which involves rewards that are sparse in two senses: they are \textit{spatially} sparse, because the agent must be close to the target by the end of an episode to receive appreciable reward; and they are \textit{temporally sparse}, because the agent is rewarded only at the end of episodes, and largely based on the result of a temporal sequence of actions rather than any individual action. We found that spatial sparsity strongly modulated learning dynamics, with small increases in the initial agent-target distance potentially yielding exponential increases in learning time. Meanwhile, we found that the task's temporal extent did not create any additional exponential slowdowns, but rather produced a trajectory refinement phase after learning to reach the target. Here, at least, temporal reward sparsity is much more benign than spatial sparsity.



\paragraph{Inductive biases due to learning dynamics.} It is well-known that learning dynamics can produce useful inductive biases. For example, noise associated with stochastic gradient descent has been observed to identify `flatter' loss landscape optima which may generalize better \citep{keskar_sharp_minima_sgd_2017,smith_bayes_sgd_2018,smith_sgdnoise_2020}, and neural networks learn `simpler' functions before more complex ones \citep{bietti_inductive_2019,canatar_spectral_2021,mingard_deep_2025}. The implicit curriculum learning effect we identified is interesting in part because it arises from gradient ascent and a simple (linear) architecture: the critical component appears to be the sparse, non-quadratic return. It is also interesting that what is ostensibly extra task complexity ($F \geq 1$ features and $D \cdot F$ parameters to learn) can make learning substantially easier rather than harder.

\paragraph{Limitations.} \textbf{We do not claim that our model explains the impact of reward sparsity in \textit{all} environments.} Rather, we only claim that our model offers an analytically tractable setting in which questions about reward sparsity are easier to pose and study. We expect that in realistic RL settings, other factors come into play. For example, in closed-loop settings, learning dynamics are more complex because previous actions affect future states and actions, which complexifies gradients.

One significant assumption we made is that policies associated with different time steps do not share parameters; this assumption makes learning dynamics approximately linear when rewards are `dense' (i.e., when $\kappa$ is small) and keeps them simple in a variety of other ways. We made this assumption because the non-convex nature of the return (Eq. \ref{eq:return}) already greatly complexifies learning dynamics, but acknowledge that learning dynamics ought to change when this assumption does not hold. Early analyses of closed versus open-loop learning dynamics have already suggested interesting differences \citep{ger_closedloop_2025}. Another strong assumption we made is that policies are parameterized by a simple linear function approximator. One expects more complex dynamics with multiple regimes---e.g., rich versus lazy \citep{domine2025from}---given deeper and possibly nonlinear networks. A particularly interesting possible future direction is to study the learning dynamics of policies parameterized by linear \textit{recurrent} networks, which are challenging to understand even in supervised settings \citep{blake_RNN_2025,proca2025learning,vastola2026rnn}.

\clearpage

\subsubsection*{Acknowledgments}

KR was funded by the NIH (RF1DA056403, U01NS136507), James S. McDonnell Foundation (220020466), Simons Foundation (Pilot Extension-00003332-02), McKnight Endowment Fund, CIFAR Azrieli Global Scholar Program, NSF (2046583), a Harvard Medical School Neurobiology Lefler Small Grant Award, and a Harvard Medical School Dean’s Innovation Award. This work has been made possible in part by a gift from the Chan Zuckerberg Initiative Foundation to establish the Kempner Institute for the Study of Natural and Artificial Intelligence at Harvard University.

\subsection*{AI use statement}

In this work, we used generative AI tools for brainstorming, literature search, to assist with checking the mathematical correctness of our results, and to assist with code generation. The authors have checked all text and results and take full responsibility for them.

%
%


\bibliography{cl2_bib}
\bibliographystyle{iclr2027_conference}

\clearpage

\appendix

\section{Derivation of average learning dynamics}
\label{app:dyn_deriv}


In this appendix, we derive the ODEs which govern the policy learning dynamics of our pursuit setting. This amounts to doing a large number of Gaussian integrals. See Sec. \ref{sec:form} for more details on how various aspects of our setting are defined.

\subsection{Review of assumptions regarding agent policy}

Recall from Sec. \ref{sec:form} that agents must learn a policy
\begin{equation} 
\pi(\vec{u}_{0:T} | \vec{z}) = \prod_{t = 0}^{M-1} \pi( \vec{u}_t | \vec{z}) \hspace{1in} \pi(\vec{u}_t | \vec{z}) = \mathcal{N}(\vec{u}_t ; \bar{\vec{u}}_t(\vec{z}), \frac{1}{\Delta t} \vec{B}_t^{-1}) 
\end{equation}
for selecting a temporal sequence of $M \geq 1$ $D$-dimensional actions $\vec{u}_0, \vec{u}_1, ..., \vec{u}_{M-1}$ given a state vector $\vec{z} \in \mathbb{R}^F$. These correspond to the agent's velocity vectors at each time step. At the end of the episode, the agent receives two rewards: one which penalizes the size of these velocities, and one which depends on the distance between their \textit{position} (or \textit{endpoint})
\begin{equation}
\vec{x}_T := \sum_{t = 0}^{M-1} \vec{u}_t \Delta t
\end{equation}
and the location $\vec{f}_T(\vec{z})$ of the target. Note that the final position of the agent is defined, as usual, as their integrated velocity vector.

The agent's policy at each time step is Gaussian with a state-dependent mean $\bar{\vec{u}}_t(\vec{z})$ and covariance matrix $\vec{B}_t^{-1}/(\Delta t)$, but only the mean action at each time step is learnable. We assume that the agent's average action depends linearly on $\vec{z}$, i.e., 
\begin{equation}
\bar{\vec{u}}_t(\vec{z}) = \frac{1}{T} \vec{b}_t + \frac{1}{T} \vec{W}_t \vec{z} \ ,
\end{equation}
and that the \textit{bias vector} $\vec{b}_t \in \mathbb{R}^D$ and \textit{weight matrix} $\vec{W}_t \in \mathbb{R}^{D \times F}$ associated with each time step's policy are learnable. The factors of $(1/T)$ are included to ensure that the bias vector and weight matrix are both unitless, even though the average action must have units of velocity (i.e., inverse time).

We are effectively in a teacher-student setting, since the target's location is assumed to depend on $\vec{z}$ as
\begin{equation}
\vec{f}_T(\vec{z}) = \vec{b}^f + \vec{W}^f \vec{z}
\end{equation}
where $\vec{b}^f \in \mathbb{R}^D$ and $\vec{W}^f \in \mathbb{R}^{D \times F}$. The average endpoint of the agent, meanwhile, equals
\begin{equation}
\bar{\vec{x}}_T(\vec{z}) := \sum_{t = 0}^{M-1} \bar{\vec{u}}_t(\vec{z}) \Delta t = \left( \sum_{t = 0}^{M-1} \vec{b}_t \frac{\Delta t}{T}  \right) + \left( \sum_{t = 0}^{M-1} \vec{W}_t \frac{\Delta t}{T} \right) \vec{z} = \vec{b} + \vec{W} \vec{z}
\end{equation}
where we have defined the overall (i.e., endpoint-associated) bias and weight matrix via
\begin{equation}
\begin{split}
\vec{b} &:= \sum_{t = 0}^{M-1} \vec{b}_t \frac{\Delta t}{T} = \frac{1}{M} \sum_{t = 0}^{M-1} \vec{b}_t  \\
\vec{W} &:= \sum_{t = 0}^{M-1} \vec{W}_t \frac{\Delta t}{T} = \frac{1}{M} \sum_{t = 0}^{M-1} \vec{W}_t  \ .
\end{split}
\end{equation}
In what follows, we study the learning dynamics of $\vec{b}_t$ and $\vec{W}_t$. We may also directly study the learning dynamics of $\vec{b}$ and $\vec{W}$, which summarize the endpoint-associated component of the agent's learned policy.

\subsection{REINFORCE policy learning}

We assume agents learn through episodic REINFORCE, whose general parameter updates for a parameter $\theta$ of the policy have the form
\begin{equation}
\Delta \theta = \eta \Delta \tau \ \nabla_{\theta} \log \pi( \vec{u}_{0:T} | \vec{z}) \ G(\vec{u}_{0:T}, \vec{z}) \ ,
\end{equation}
where $\Delta \theta$ is the size of the update after an episode, $G(\vec{u}_{0:T}, \vec{z})$ is the return associated with that episode, and $\eta (\Delta \tau) > 0$ is the learning rate.

We make two assumptions which greatly simplify learning dynamics. First, we assume that the total number of episodes ($N_E$, say) is taken to be extremely large, and the learning rate is chosen to scale inversely with $N_E$, i.e., $\Delta \tau := 1/N_E$. 

This assumption is significant because it allows us to eliminate all randomness associated with parameter updates. Normally, parameter updates would be stochastic due to randomness in the policy and state vector draws (since $\vec{z} \sim \mathcal{N}(\vec{0}, \vec{\Sigma})$). But variance in $\Delta \theta$ scales like
\begin{equation}
\text{var}(\Delta \theta) \sim (\Delta \tau)^2 \ ,
\end{equation}
which is too small in the $N_E \to \infty$ limit to contribute to learning. Since there is no variance, we obtain that updates correspond precisely to the \textit{mean update}, i.e., 
\begin{equation}
\Delta \theta = \eta \Delta \tau \ \mathbb{E}_{\vec{u}_{0:T}, \vec{z}}\left[ \nabla_{\theta} \log \pi( \vec{u}_{0:T} | \vec{z}) \ G(\vec{u}_{0:T}, \vec{z}) \right] \ .
\end{equation}
In the $\Delta \tau \to 0$ limit, we recover an ODE
\begin{equation} \label{eq:REINFORCE_ODE_general}
\frac{d}{d\tau} \theta = \eta \ \mathbb{E}_{\vec{u}_{0:T}, \vec{z}}\left[ \nabla_{\theta} \log \pi( \vec{u}_{0:T} | \vec{z}) \ G(\vec{u}_{0:T}, \vec{z}) \right]
\end{equation}
where we use $\tau \geq 0$ to parameterize `learning time'. In the limit, the number of episodes is so large that it is effectively treated as a continuous variable. It is for this reason that we refer to this as the ``continuously-many episodes limit''. Since $\Delta \tau$ no longer appears, we will now refer to $\eta > 0$ as the learning rate.


We must now specialize Eq. \ref{eq:REINFORCE_ODE_general} to our particular choice of policy. To ensure appropriate scaling of the learning dynamics with the number of time steps $M$, we will take
\begin{equation}
\eta \to \eta \frac{T}{\Delta t} = \eta M \ .
\end{equation}
Now if $\theta_t$ is a parameter of the time step $t$ mean action $\bar{\vec{u}}_t(\vec{z})$, then
\begin{equation}
\frac{\partial}{\partial \theta_t}  \log \pi( \vec{u}_{0:T} | \vec{z}) = \frac{\partial \bar{\vec{u}}_t(\vec{z})^T}{\partial \theta_t} \vec{B}_t \Delta t (\vec{u}_t - \bar{\vec{u}}_t(\vec{z})) \ ,
\end{equation}
which implies that
\begin{equation}
\frac{d}{d\tau} \theta_t = \frac{\eta T}{\Delta t}   \ \mathbb{E}_{\vec{z}}\left\{ \frac{\partial \bar{\vec{u}}_t(\vec{z})^T}{\partial \theta_t} \vec{B}_t \Delta t \ \mathbb{E}_{\vec{u}_{0:T} | \vec{z}}\left[ \ (\vec{u}_t - \bar{\vec{u}}_t(\vec{z})) \ G(\vec{u}_{0:T}, \vec{z}) \right]  \right\} \ .
\end{equation}
This is the general ODE which appears in Sec. \ref{sec:form}. We can specialize it further to obtain learning dynamics for the biases $\vec{b}_t$ and weights $\vec{W}_t$. Since
\begin{equation}
\begin{split}
\frac{\partial \bar{u}_{t k}(\vec{z})}{\partial b_{t' j}} &=  \frac{1}{T} \delta_{k j}  \delta_{t t'}  \\
\frac{\partial \bar{u}_{t k}(\vec{z})}{\partial W_{t', ij}} &=  \frac{1}{T} \delta_{k i} z_j \delta_{t t'}  \ ,
\end{split}
\end{equation}
we have
\begin{equation}
\begin{split}
\frac{d}{d\tau} \vec{b}_t &= \eta \vec{B}_t  \ \mathbb{E}_{\vec{z}}\left\{  \ \mathbb{E}_{\vec{u}_{0:T} | \vec{z}}\left[ \ (\vec{u}_t - \bar{\vec{u}}_t(\vec{z})) \ G(\vec{u}_{0:T}, \vec{z}) \right]  \right\} \\
\frac{d}{d\tau} \vec{W}_t &= \eta \vec{B}_t  \ \mathbb{E}_{\vec{z}}\left\{  \ \mathbb{E}_{\vec{u}_{0:T} | \vec{z}}\left[ \ (\vec{u}_t - \bar{\vec{u}}_t(\vec{z})) \ G(\vec{u}_{0:T}, \vec{z}) \right] \ \vec{z}^T  \right\} \ .
\end{split}
\end{equation}
All that remains to do is to explicitly perform the averages over actions $\vec{u}_{0:T}$ and states $\vec{z}$.


\subsection{Average over actions}

In each parameter update, the same average
\begin{equation}
A = \mathbb{E}_{\vec{u}_{0:T} | \vec{z}}\left[ \ (\vec{u}_t - \bar{\vec{u}}_t(\vec{z})) \ G(\vec{u}_{0:T}, \vec{z}) \right] 
\end{equation}
over actions $\vec{u}_{0:T}$ appears. In this subsection, we compute it explicitly. Recall that the return associated with our setting is (see Sec. \ref{sec:form}) 
\begin{equation} \label{eq:return_app}
G( \vec{u}_{0:T}, \vec{z} ) = G_{\text{effort}} + G_{\text{target}} =  \sum_{t=0}^{M-1} -\frac{\alpha}{2} \Vert \vec{u}_t \Vert_2^2   \ \Delta t +  g \left( \frac{\kappa}{2 \pi} \right)^{D/2} e^{- \frac{\kappa}{2} \Vert \vec{x}_T - \vec{f}_T(\vec{z}) \Vert_2^2}  
\end{equation}
where $\alpha > 0$ controls the effort cost, $g > 0$ controls the size of the target-associated reward, and $\kappa > 0$ controls the sparsity of that reward. We can hence break up the overall average into two parts:
\begin{equation}
\begin{split}
A_{\text{effort}} &:=- \frac{\alpha}{2} \sum_{t' = 0}^{M-1} \Delta t \ \mathbb{E}_{\vec{u}_{0:T} | \vec{z}}\left[ \ (\vec{u}_t - \bar{\vec{u}}_t(\vec{z})) \ \Vert \vec{u}_{t'} \Vert_2^2 \right] \\
A_{\text{target}} &:= g \ \mathbb{E}_{\vec{u}_{0:T} | \vec{z}}\left[ \ (\vec{u}_t - \bar{\vec{u}}_t(\vec{z})) \ \left( \frac{\kappa}{2 \pi} \right)^{D/2} e^{- \frac{\kappa}{2} \Vert \vec{x}_T - \vec{f}_T(\vec{z}) \Vert_2^2}  \right] \ .
\end{split}
\end{equation}

\paragraph{Effort term.} Evaluating the first average is easy. First, since the $\vec{u}_t$ are independent of one another, note that the average vanishes unless $t' = t$. Then we have
\begin{equation}
\begin{split}
A_{\text{effort}} &= - \frac{\alpha}{2} \Delta t \ \mathbb{E}_{\vec{u}_{0:T} | \vec{z}}\left[ \ (\vec{u}_t - \bar{\vec{u}}_t) \ \Vert \vec{u}_{t} \Vert_2^2 \right] \\
&= - \frac{\alpha}{2} \Delta t \ \mathbb{E}_{\vec{u}_{0:T} | \vec{z}}\left[ \ (\vec{u}_t - \bar{\vec{u}}_t) \ \Vert \vec{u}_{t} - \bar{\vec{u}}_t + \bar{\vec{u}}_t \Vert_2^2 \right] \\
&= - \frac{\alpha}{2} \Delta t \ \mathbb{E}_{\vec{u}_{0:T} | \vec{z}}\left\{ \ (\vec{u}_t - \bar{\vec{u}}_t) \left[ \Vert \vec{u}_t - \bar{\vec{u}}_t \Vert_2^2 + 2 ( \vec{u}_t - \bar{\vec{u}}_t )^T \bar{\vec{u}}_t + \Vert \bar{\vec{u}}_t \Vert_2^2 \right] \ \right\} \\
&= - \alpha \Delta t \ \mathbb{E}_{\vec{u}_{0:T} | \vec{z}}\left[  (\vec{u}_t - \bar{\vec{u}}_t)  ( \vec{u}_t - \bar{\vec{u}}_t )^T \right] \bar{\vec{u}}_t  \\
&= - \alpha \Delta t \ \frac{1}{\Delta t} \vec{B}_t^{-1} \bar{\vec{u}}_t  \\
&= - \alpha \vec{B}_t^{-1} \ \bar{\vec{u}}_t  \ .
\end{split}
\end{equation}

\paragraph{Target-associated term.} The target-associated term is significantly more difficult to treat. To deal with it, we will use several tricks. First, we will introduce the function
\begin{equation}
I(\vec{J}_{0:T}) := \mathbb{E}_{\vec{u}_{0:T}}\left[ \left( \frac{\kappa}{2 \pi} \right)^{D/2} e^{- \frac{\kappa}{2} \Vert \vec{x}_T - \vec{f}_T \Vert_2^2 + \sum_t \vec{J}_t^T (\vec{u}_t - \bar{\vec{u}}_t) \Delta t}     \right] \ ,
\end{equation}
where the vectors $\vec{J}_0$, $\vec{J}_1$, ..., $\vec{J}_{M-1}$ are `source terms'. Importantly, the average we care about corresponds to the derivative of $I$ with respect to a $\vec{J}_t$ evaluated where all source terms are zero:
\begin{equation}
\frac{1}{\Delta t} \left. \frac{\partial I}{\partial \vec{J}_t} \right|_{\vec{J}_t = \vec{0} \ \forall t} =  \mathbb{E}_{\vec{u}_{0:T}}\left[ ( \vec{u}_t - \bar{\vec{u}}_t) \left( \frac{\kappa}{2 \pi} \right)^{D/2} e^{- \frac{\kappa}{2} \Vert \vec{x}_T - \vec{f}_T \Vert_2^2}     \right]  \ .
\end{equation}
Hence, it suffices to compute $I(\vec{J}_{0:T})$ and then take the appropriate derivative.

Second, we will exploit a useful integral representation of the target-associated reward which comes from the characteristic function of a multivariate Gaussian. We have that
\begin{equation}
\begin{split}
\left( \frac{\kappa}{2 \pi} \right)^{D/2} e^{- \frac{\kappa}{2} \Vert \vec{x}_T - \vec{f}_T \Vert_2^2}   &= \int \frac{d\vec{p}}{(2 \pi)^D} e^{- i\vec{p}^T (\vec{x}_T - \vec{f}_T) - \frac{1}{2 \kappa} \vec{p}^T \vec{p}} \\
&= \int \frac{d\vec{p}}{(2 \pi)^D} e^{- i\vec{p}^T \left( \sum_t \vec{u}_t \Delta t - \vec{f}_T \right) - \frac{1}{2 \kappa} \vec{p}^T \vec{p}} \ .
\end{split}
\end{equation}
Using this integral representation, $I(\vec{J}_{0:T})$ can be written
\begin{equation}
\begin{split}
I(\vec{J}_{0:T}) &= \int \frac{d\vec{p}}{(2 \pi)^D} e^{ i \vec{p}^T \vec{f}_T - \frac{1}{2 \kappa} \vec{p}^T \vec{p}} \ \mathbb{E}_{\vec{u}_{0:T}}\left[  e^{ - i\vec{p}^T \left( \sum_t \vec{u}_t \Delta t  \right)  + \sum_t \vec{J}_t^T (\vec{u}_t - \bar{\vec{u}}_t) \Delta t}     \right] \\
&=  \int \frac{d\vec{p}}{(2 \pi)^D} e^{i \vec{p}^T \vec{f}_T - \frac{1}{2 \kappa} \vec{p}^T \vec{p} - \sum_t \vec{J}_t^T \bar{\vec{u}}_t \Delta t} \ \mathbb{E}_{\vec{u}_{0:T}}\left[  e^{\sum_t \left( \vec{J}_t - i \vec{p}  \right)^T \vec{u}_t \Delta t}     \right] \\
&=  \int \frac{d\vec{p}}{(2 \pi)^D} e^{i \vec{p}^T \vec{f}_T - \frac{1}{2 \kappa} \vec{p}^T \vec{p} - \sum_t \vec{J}_t^T \bar{\vec{u}}_t \Delta t + \sum_t \left( \vec{J}_t - i \vec{p}  \right)^T \bar{\vec{u}}_t \Delta t + \frac{\Delta t}{2} ( \vec{J}_t - i \vec{p}  )^T \vec{B}^{-1}_t ( \vec{J}_t - i \vec{p}  )}   \ .
\end{split}
\end{equation}
The argument of the exponential can be rewritten as
\begin{equation}
\left( \sum_t \frac{\Delta t}{2} \vec{J}_t^T \vec{B}_t^{-1} \vec{J}_t \right) - i \vec{p}^T \left( \sum_t ( \bar{\vec{u}}_t + \vec{B}_t^{-1} \vec{J}_t ) \Delta t - \vec{f}_T    \right) - \frac{1}{2} \vec{p}^T \left[ \frac{1}{\kappa} \vec{I} + \sum_t \vec{B}_t^{-1} \Delta t  \right] \vec{p} \ .
\end{equation}
To ease notation, define the $D \times D$ positive definite matrix 
\begin{equation}
\boxed{ \ \tilde{\vec{B}}^{-1} := \frac{1}{\kappa} \vec{I} + \sum_t \vec{B}_t^{-1} \Delta t \ } \ .
\end{equation}
Then $I(\vec{J}_{0:T})$ equals
\begin{equation}
\begin{split}
I(\vec{J}_{0:T}) &= e^{\sum_t \frac{\Delta t}{2} \vec{J}_t^T \vec{B}_t^{-1} \vec{J}_t} \int \frac{d\vec{p}}{(2 \pi)^D} e^{- i \vec{p}^T \left( \sum_t ( \bar{\vec{u}}_t + \vec{B}_t^{-1} \vec{J}_t ) \Delta t - \vec{f}_T    \right) - \frac{1}{2} \vec{p}^T \tilde{\vec{B}}^{-1} \vec{p}}   \\
&= e^{\sum_t \frac{\Delta t}{2} \vec{J}_t^T \vec{B}_t^{-1} \vec{J}_t} \sqrt{\frac{\det\tilde{\vec{B}}}{(2 \pi)^D }} e^{- \frac{1}{2} \left[  \sum_t ( \bar{\vec{u}}_t + \vec{B}_t^{-1} \vec{J}_t ) \Delta t - \vec{f}_T  \right]^T \tilde{\vec{B}} \left[  \sum_t ( \bar{\vec{u}}_t + \vec{B}_t^{-1} \vec{J}_t ) \Delta t - \vec{f}_T  \right]} \ .
\end{split}
\end{equation}
Now that we have evaluated $I(\vec{J}_{0:T})$, we can take its derivative. It equals
\begin{equation}
\begin{split}
\frac{\partial I}{\partial \vec{J}_t} &= \Delta t \left[  \vec{B}_t^{-1} \vec{J}_t  - \vec{B}_t^{-1} \tilde{\vec{B}} \left(  \sum_t ( \bar{\vec{u}}_t + \vec{B}_t^{-1} \vec{J}_t ) \Delta t - \vec{f}_T  \right)   \right] I(\vec{J}_{0:T}) \ ,
\end{split}
\end{equation}
and when we set all $\vec{J}_t = \vec{0}$, we find
\begin{equation}
\begin{split}
\frac{1}{\Delta t} \left. \frac{\partial I}{\partial \vec{J}_t} \right|_{\vec{J}_t = \vec{0} \ \forall t} &=  \vec{B}_t^{-1} \tilde{\vec{B}} \left(  \vec{f}_T - \sum_t  \bar{\vec{u}}_t \ \Delta t  \right)    I(\vec{J}_{0:T} = \vec{0}) \\
&=  \vec{B}_t^{-1} \tilde{\vec{B}} \left(  \vec{f}_T - \sum_t  \bar{\vec{u}}_t \ \Delta t  \right)  \sqrt{\frac{\det\tilde{\vec{B}}}{(2 \pi)^D }} e^{- \frac{1}{2} (  \sum_t \bar{\vec{u}}_t  \Delta t - \vec{f}_T  )^T \tilde{\vec{B}} (  \sum_t  \bar{\vec{u}}_t  \Delta t - \vec{f}_T  )} \ .
\end{split}
\end{equation}
Recalling that the average learned endpoint is $\bar{\vec{x}}_T := \sum_t \bar{\vec{u}}_t \Delta t$, we conclude that
\begin{equation}
\boxed{ \ A_{\text{target}} = g \vec{B}_t^{-1} \tilde{\vec{B}} \left(  \vec{f}_T - \bar{\vec{x}}_T  \right)  \sqrt{\frac{\det\tilde{\vec{B}}}{(2 \pi)^D }} e^{- \frac{1}{2} (  \bar{\vec{x}}_T - \vec{f}_T  )^T \tilde{\vec{B}} (  \bar{\vec{x}}_T - \vec{f}_T  )} \ } \ .
\end{equation}
Overall,
\begin{equation}
\boxed{ \ \mathbb{E}_{\vec{u}_{0:T} | \vec{z}}\left[  (\vec{u}_t - \bar{\vec{u}}_t(\vec{z})) \ G(\vec{u}_{0:T}, \vec{z}) \right] = - \alpha \vec{B}_t^{-1} \ \bar{\vec{u}}_t  +   g \vec{B}_t^{-1} \tilde{\vec{B}} \left(  \vec{f}_T - \bar{\vec{x}}_T  \right)  \sqrt{\frac{\det\tilde{\vec{B}}}{(2 \pi)^D }} e^{- \frac{1}{2} (  \bar{\vec{x}}_T - \vec{f}_T  )^T \tilde{\vec{B}} (  \bar{\vec{x}}_T - \vec{f}_T  )} \ } \ .
\end{equation}
This implies that our REINFORCE ODEs have the general form
\begin{equation} \label{eq:REINFORCE_ODE_beforez}
\boxed{ \ \frac{d}{d\tau} \theta_t = \eta T \ \mathbb{E}_{\vec{z}}\left\{ \frac{\partial \bar{\vec{u}}_t(\vec{z})^T}{\partial \theta_t}  \left[ - \alpha \bar{\vec{u}}_t  +   g \tilde{\vec{B}} \left(  \vec{f}_T - \bar{\vec{x}}_T  \right)  \sqrt{\frac{\det\tilde{\vec{B}}}{(2 \pi)^D }} e^{- \frac{1}{2} (  \bar{\vec{x}}_T - \vec{f}_T  )^T \tilde{\vec{B}} (  \bar{\vec{x}}_T - \vec{f}_T  )} \right]  \right\} \ } \ .
\end{equation}
Specializing to $\vec{b}_t$ and $\vec{W}_t$, we obtain
\begin{equation} \label{eq:bW_before}
\begin{split}
\frac{d}{d\tau} \vec{b}_t &= \eta \ \mathbb{E}_{\vec{z}}\left\{  - \alpha \bar{\vec{u}}_t  +   g \tilde{\vec{B}} \left(  \vec{f}_T - \bar{\vec{x}}_T  \right)  \sqrt{\frac{\det\tilde{\vec{B}}}{(2 \pi)^D }} e^{- \frac{1}{2} (  \bar{\vec{x}}_T - \vec{f}_T  )^T \tilde{\vec{B}} (  \bar{\vec{x}}_T - \vec{f}_T  )}  \right\} \\
\frac{d}{d\tau} \vec{W}_t &= \eta \ \mathbb{E}_{\vec{z}}\left\{  \left[ - \alpha \bar{\vec{u}}_t  +   g \tilde{\vec{B}} \left(  \vec{f}_T - \bar{\vec{x}}_T  \right)  \sqrt{\frac{\det\tilde{\vec{B}}}{(2 \pi)^D }} e^{- \frac{1}{2} (  \bar{\vec{x}}_T - \vec{f}_T  )^T \tilde{\vec{B}} (  \bar{\vec{x}}_T - \vec{f}_T  )} \right]  \vec{z}^T \right\}  \ .
\end{split}
\end{equation}

\subsection{The effective return}

A useful feature of our dynamics is that they correspond to gradient ascent up the \textit{effective return}
\begin{equation}
G_{\text{eff}} := \mathbb{E}_{\vec{z}}\left[ - \sum_t \frac{\alpha}{2} \Vert \bar{\vec{u}}_t \Vert_2^2 \Delta t  + g \sqrt{\frac{\det \tilde{\vec{B}}}{(2 \pi)^D}} \ e^{- \frac{1}{2} \left(   \bar{\vec{x}}_T   - \vec{f}_T \right)^T \tilde{\vec{B}} \left(  \bar{\vec{x}}_T  - \vec{f}_T  \right) }   \right] \ ,
\end{equation}
which only depends on policy parameters rather than on any specific states and actions. In terms of this quantity, the general REINFORCE ODE (Eq. \ref{eq:REINFORCE_ODE_beforez})  becomes
\begin{equation}
\frac{d}{d\tau} \theta(t) = \frac{\eta T}{\Delta t} \ \frac{\partial G_{\text{eff}}}{\partial \theta(t)} \ ,
\end{equation} 
and $\vec{b}_t$ and $\vec{W}_t$ dynamics (Eq. \ref{eq:bW_before}) become
\begin{equation}
\begin{split}
\frac{d}{d\tau} \vec{b}_t &= \frac{\eta T}{\Delta t} \ \frac{\partial G_{\text{eff}}}{\partial \vec{b}_t} \\
\frac{d}{d\tau} \vec{W}_t &= \frac{\eta T}{\Delta t} \ \frac{\partial G_{\text{eff}}}{\partial \vec{W}_t} \ .
\end{split}
\end{equation} 
Simplifying Eq. \ref{eq:bW_before} would ordinarily require taking multiple expectations over $\vec{z}$, since slightly different expectations appear in $\vec{b}_t$ and $\vec{W}_t$ dynamics. But this fact allows us to simultaneously derive both by first taking a \textit{single} expectation over $\vec{z}$ (to compute $G_{\text{eff}}$), and then taking the appropriate derivatives.



\subsection{Pure bias learning dynamics}

We stop here to note that we already have our final result for pure bias learning, i.e., the case where there are no features and no feature-associated weights. The mean action at time step $t$ is simply $\bar{\vec{u}}_t = \vec{b}_t/T$, the average endpoint is $\bar{\vec{x}}_T = \sum_t \vec{b}_t \frac{\Delta t}{T} = \vec{b}$, the target location is fixed at $\vec{f}_T = \vec{b}^f$, and learning dynamics read
\begin{equation} 
\begin{split}
\frac{d}{d\tau} \vec{b}_t &= \eta \ \left\{  - \alpha \bar{\vec{u}}_t  +   g \tilde{\vec{B}} \left(  \vec{f}_T - \bar{\vec{x}}_T  \right)  \sqrt{\frac{\det\tilde{\vec{B}}}{(2 \pi)^D }} e^{- \frac{1}{2} (  \bar{\vec{x}}_T - \vec{f}_T  )^T \tilde{\vec{B}} (  \bar{\vec{x}}_T - \vec{f}_T  )}  \right\} \\
&= \eta \ \left\{  - \frac{\alpha}{T} \vec{b}_t  +   g \tilde{\vec{B}} \left(  \vec{b}^f - \vec{b} \right)  \sqrt{\frac{\det\tilde{\vec{B}}}{(2 \pi)^D }} e^{- \frac{1}{2} (  \vec{b} - \vec{b}^f  )^T \tilde{\vec{B}} (  \vec{b} - \vec{b}^f  )}  \right\} \ .
\end{split}
\end{equation}
This follows immediately from Eq. \ref{eq:bW_before}, since we can discard the expectation over $\vec{z}$. 

We can suggestively rewrite these dynamics by defining the \textit{hitting probability}
\begin{equation}
\boxed{ \ p_{\text{hit}}(\vec{b}) := \sqrt{\frac{\det\tilde{\vec{B}}}{(2 \pi)^D }} e^{- \frac{1}{2} (  \vec{b} - \vec{b}^f  )^T \tilde{\vec{B}} (  \vec{b} - \vec{b}^f  )} \ } \ .
\end{equation}
In terms of it, we have
\begin{equation} 
\boxed{ \ \frac{d}{d\tau} \vec{b}_t = \eta \ \left\{  - \frac{\alpha}{T} \vec{b}_t  +   g \tilde{\vec{B}} \left(  \vec{b}^f - \vec{b} \right)  p_{\text{hit}}(\vec{b}) \right\} \ } \ .
\end{equation}
By averaging the ODEs associated with each $\vec{b}_t$, we can obtain an ODE which directly tracks the time evolution of the endpoint-associated bias $\vec{b} = \sum_t \vec{b}_t \frac{\Delta t}{T}$:
\begin{equation} 
\boxed{ \ \frac{d}{d\tau} \vec{b} = \eta \ \left\{  - \frac{\alpha}{T} \vec{b}  +   g \tilde{\vec{B}} \left(  \vec{b}^f - \vec{b} \right)  p_{\text{hit}}(\vec{b}) \right\} \ } \ .
\end{equation}
Notice that it is the same as the ODE for the individual $\vec{b}_t$ except for the effort term.

\subsection{Average over states}

We will now determine the effect of state variability on learning by averaging updates over the observed state vector, which is distributed according to $\vec{z} \sim \mathcal{N}(\vec{0}, \vec{\Sigma})$, where $\vec{\Sigma}$ is a $F \times F$ covariance matrix.

We can do this by computing the effective return
\begin{equation}
G_{\text{eff}} := \mathbb{E}_{\vec{z}}\left[ - \sum_t \frac{\alpha}{2} \Vert \bar{\vec{u}}_t \Vert_2^2 \Delta t  + g \sqrt{\frac{\det \tilde{\vec{B}}}{(2 \pi)^D}} \ e^{- \frac{1}{2} \left(   \bar{\vec{x}}_T   - \vec{f}_T \right)^T \tilde{\vec{B}} \left(  \bar{\vec{x}}_T  - \vec{f}_T  \right) }   \right] \ ,
\end{equation}
and then taking derivatives with respect to $\vec{b}_t$ and $\vec{W}_t$. As when we took the average over actions, the average we must do involves two terms: an effort-associated term, and a target-associated term. More precisely, $G_{\text{eff}} = A_{\text{effort}} + A_{\text{target}}$, where
\begin{equation}
\begin{split}
A_{\text{effort}} &:= - \frac{\alpha \Delta t}{2} \sum_t \mathbb{E}_{\vec{z}}\left[\Vert \bar{\vec{u}}_t \Vert_2^2 \right] \\
A_{\text{target}} &:= g \ \mathbb{E}_{\vec{z}}\left[  \sqrt{\frac{\det \tilde{\vec{B}}}{(2 \pi)^D}} \ e^{- \frac{1}{2} \left(   \bar{\vec{x}}_T   - \vec{f}_T \right)^T \tilde{\vec{B}} \left(  \bar{\vec{x}}_T  - \vec{f}_T  \right) }  \right]  \ .
\end{split}
\end{equation}

\paragraph{Effort term.} Once again, evaluating the first average is easy. We have
\begin{equation}
\begin{split}
A_{\text{effort}} &= - \frac{\alpha \Delta t}{2 T^2} \sum_t \mathbb{E}_{\vec{z}}\left[  (\vec{b}_t + \vec{W}_t \vec{z})^T  (\vec{b}_t + \vec{W}_t \vec{z})   \right] \\
&= - \frac{\alpha \Delta t}{2 T^2} \sum_t \mathbb{E}_{\vec{z}}\left[ \Vert \vec{b}_t \Vert_2^2 + 2 \vec{b}_t^T \vec{W}_t \vec{z} + \vec{z}^T \vec{W}_t^T \vec{W}_t \vec{z}   \right] \\
&= - \frac{\alpha \Delta t}{2 T^2} \sum_t \left( \ \Vert \vec{b}_t \Vert_2^2  + \text{tr}[ \vec{\Sigma} \vec{W}_t^T \vec{W}_t  ] \ \right) \ .
\end{split}
\end{equation}

\paragraph{Target-associated term.} Once again, evaluating the other term is significantly more complicated. 
Since
\begin{equation}
\bar{\vec{x}}_T - \vec{f}_T = \vec{b} - \vec{b}^f + (\vec{W} - \vec{W}^f) \vec{z} = \Delta \vec{b} + \Delta \vec{W} \vec{z} \ ,
\end{equation}
where we define the bias and weight discrepancies $\Delta \vec{b} := \vec{b} - \vec{b}^f$ and $\Delta \vec{W} := \vec{W} - \vec{W}^f$, the integral of interest is
\begin{equation}
\begin{split}
A_{\text{target}} &= g \ \mathbb{E}_{\vec{z}}\left[  \sqrt{\frac{\det \tilde{\vec{B}}}{(2 \pi)^D}} \ e^{- \frac{1}{2} \left(  \Delta \vec{b} + \Delta \vec{W} \vec{z} \right)^T \tilde{\vec{B}} \left(  \Delta \vec{b} + \Delta \vec{W} \vec{z}  \right) }  \right]  \ .
\end{split}
\end{equation}
We will once again exploit a useful integral representation of a multivariate Gaussian to rewrite the reward-associated Gaussian:
\begin{equation}
\sqrt{\frac{\det \tilde{\vec{B}}}{(2 \pi)^D}} e^{- \frac{1}{2} \left[ \Delta \vec{W} \vec{z} + \Delta \vec{b} \right]^T \tilde{\vec{B}} \left[ \Delta \vec{W} \vec{z} + \Delta \vec{b}  \right] }  = \int \frac{d\vec{p}}{(2 \pi)^D} e^{- i \vec{p}^T (\Delta \vec{b} + \Delta \vec{W} \vec{z}) - \frac{1}{2} \vec{p}^T \tilde{\vec{B}}^{-1} \vec{p}} \ .
\end{equation}
Using this integral representation, $A_{\text{target}}$ becomes
\begin{equation}
\begin{split}
A_{\text{target}} &= g \int \frac{d\vec{p}}{(2 \pi)^D} e^{- i \vec{p}^T \Delta \vec{b}  - \frac{1}{2} \vec{p}^T \tilde{\vec{B}}^{-1} \vec{p}} \ \mathbb{E}_{\vec{z}}\left[ e^{- i \vec{p}^T \Delta \vec{W} \vec{z}} \right] \\
&= g \int \frac{d\vec{p}}{(2 \pi)^D} e^{- i \vec{p}^T \Delta \vec{b}  - \frac{1}{2} \vec{p}^T \tilde{\vec{B}}^{-1} \vec{p}} \  e^{-\frac{1}{2} \vec{p}^T (\Delta \vec{W}) \vec{\Sigma} (\Delta \vec{W})^T \vec{p} }  \\
&= g \int \frac{d\vec{p}}{(2 \pi)^D} e^{- i \vec{p}^T \Delta \vec{b}  - \frac{1}{2} \vec{p}^T \left[ \tilde{\vec{B}}^{-1}  +  (\Delta \vec{W}) \vec{\Sigma} (\Delta \vec{W})^T \right] \vec{p}}    \ .
\end{split}
\end{equation}
Performing the (Gaussian) integral, we conclude that
\begin{equation}
\boxed{ \ A_{\text{target}} =  g \sqrt{\frac{1}{(2 \pi)^D \det \vec{V}}} e^{- \frac{1}{2} (\Delta \vec{b})^T \vec{V}^{-1} (\Delta \vec{b})} \ } \ ,
\end{equation}
where we define the $D \times D$ covariance matrix (or \textit{effective target size})
\begin{equation}
\vec{V}(\vec{W}) := \tilde{\vec{B}}^{-1}  +  (\Delta \vec{W}) \vec{\Sigma} (\Delta \vec{W})^T  \ .
\end{equation}
This matrix plays a role analogous to the one $\tilde{\vec{B}}^{-1}$ plays in the case of pure bias learning. Unlike that matrix, this one depends on learned parameters: it is \textit{larger} along directions where $\vec{W}$ is very different from the target weight matrix $\vec{W}^f$.

In any case, we conclude that the effective return is
\begin{equation}
\boxed{ \ G_{\text{eff}} =   - \sum_t \frac{\alpha \Delta t}{2 T^2} \left\{ \ \Vert \vec{b}_t \Vert_2^2 + \text{tr}\left[ \vec{\Sigma} \vec{W}_t^T  \vec{W}_t  \right]  \  \right\}   + g \sqrt{\frac{1}{(2 \pi)^D \det \vec{V}}} e^{- \frac{1}{2} (\Delta \vec{b})^T \vec{V}^{-1} (\Delta \vec{b})} \ } \ .
\end{equation}

\paragraph{Derivatives of the effective return.} In order to obtain the final form of $\vec{b}_t$ and $\vec{W}_t$ dynamics, we must differentiate the effective return with respect to $\vec{b}_t$ and $\vec{W}_t$. 

The derivative with respect to $\vec{b}_t$ is simple:
\begin{equation}
\frac{\partial G_{\text{eff}}}{\partial \vec{b}_t} =   - \frac{\alpha \Delta t}{T^2} \vec{b}_t    + \frac{g \Delta t}{T} \vec{V}^{-1} ( \vec{b}^f - \vec{b}) \sqrt{\frac{1}{(2 \pi)^D \det \vec{V}}} e^{- \frac{1}{2} (\Delta \vec{b})^T \vec{V}^{-1} (\Delta \vec{b})}  \ .
\end{equation}
The derivative with respect to $\vec{W}_t$ is more complex, since we must differentiate a matrix determinant and inverse with respect to $\vec{W}_t$. Computing matrix derivatives using standard identities, we find
\begin{equation}
\begin{split}
\frac{\partial}{\partial \vec{W}} \frac{1}{2} \log\det\vec{V} &=  \vec{V}^{-1} (\vec{W} - \vec{W}^f) \vec{\Sigma} \\
\frac{\partial}{\partial \vec{W}} \left[ \frac{1}{2} (\Delta \vec{b})^T \vec{V}^{-1} (\Delta \vec{b}) \right] &= - \vec{V}^{-1} (\Delta \vec{b}) (\Delta \vec{b})^T \vec{V}^{-1} (\vec{W} - \vec{W}^f) \vec{\Sigma} \ .
\end{split}
\end{equation}
This implies, since
\begin{equation}
\log A_{\text{target}} = - \frac{1}{2} \log\det\vec{V} - \frac{1}{2} (\Delta \vec{b})^T \vec{V}^{-1} (\Delta \vec{b}) + \text{const.} \ ,
\end{equation}
we have
\begin{equation}
\begin{split}
\frac{\partial}{\partial \vec{W}} \log A_{\text{target}} &= - \vec{V}^{-1} (\vec{W} - \vec{W}^f) \vec{\Sigma} +  \vec{V}^{-1} (\Delta \vec{b}) (\Delta \vec{b})^T \vec{V}^{-1} (\vec{W} - \vec{W}^f) \vec{\Sigma}  \\
&= \left[  \vec{I} -   \vec{V}^{-1}  (\Delta \vec{b}) (\Delta \vec{b})^T  \right]  \vec{V}^{-1}  (\vec{W}^f - \vec{W}) \vec{\Sigma} \ .
\end{split}
\end{equation}
Now we can easily compute that
\begin{equation}
\begin{split}
\frac{\partial G_{\text{eff}}}{\partial \vec{W}_t}  &= - \sum_{t'} \frac{\alpha \Delta t}{2 T^2 } \frac{\partial}{\partial \vec{W}_t} \left( \text{tr}\left[ \vec{\Sigma} \vec{W}_{t'}^T  \vec{W}_{t'}  \right]  \right) +  \frac{\partial}{\partial \vec{W}_t} \left[  g \sqrt{\frac{1}{(2 \pi)^D \det \vec{V}}} e^{- \frac{1}{2} (\Delta \vec{b})^T \vec{V}^{-1} (\Delta \vec{b})}  \right]  \\
&= - \frac{\alpha \Delta t}{T^2 } \vec{W}_t \vec{\Sigma} + \frac{g \Delta t}{T} \frac{\partial}{\partial \vec{W}}\left( \log A_{\text{target}} \right) \  \sqrt{\frac{1}{(2 \pi)^D \det \vec{V}}} e^{- \frac{1}{2} (\Delta \vec{b})^T \vec{V}^{-1} (\Delta \vec{b})}   \\
&= - \frac{\alpha \Delta t}{T^2 } \vec{W}_t \vec{\Sigma} + \frac{g \Delta t}{T} \left[  \vec{I} -   \vec{V}^{-1}  (\Delta \vec{b}) (\Delta \vec{b})^T  \right]  \vec{V}^{-1}  (\vec{W}^f - \vec{W}) \vec{\Sigma} \  \frac{e^{- \frac{1}{2} (\Delta \vec{b})^T \vec{V}^{-1} (\Delta \vec{b})}}{\sqrt{(2 \pi)^D \det \vec{V}}}    \ .
\end{split}
\end{equation}

\paragraph{Final result.} Since the dynamics of $\vec{b}_t$ and $\vec{W}_t$ correspond to gradient ascent up the effective return, we obtain final ODEs
\begin{equation*}
\begin{split}
\frac{d}{d\tau} \vec{b}_t &= \eta \left\{  - \frac{\alpha}{T}  \vec{b}_t    + g \vec{V}^{-1} ( \vec{b}^f - \vec{b}) \frac{e^{- \frac{1}{2} (\Delta \vec{b})^T \vec{V}^{-1} (\Delta \vec{b})}  }{\sqrt{(2 \pi)^D \det \vec{V}}} \right\} \\
\frac{d}{d\tau} \vec{W}_t &= \eta \left\{   - \frac{\alpha}{T}  \vec{W}_t \vec{\Sigma}   + g \left[    \vec{I} - \vec{V}^{-1} (\Delta \vec{b}) (\Delta \vec{b})^T  \right] \vec{V}^{-1} (\vec{W}^f - \vec{W}) \vec{\Sigma}  \ \frac{e^{- \frac{1}{2} (\Delta \vec{b})^T \vec{V}^{-1} (\Delta \vec{b})} }{\sqrt{(2 \pi)^D \det \vec{V}}}    \right\} \ .
\end{split}
\end{equation*}
Notice that the target-associated term does not depend on $t$. This implies that the endpoint-associated quantities $\vec{b} = \sum_t \vec{b}_t \frac{\Delta t}{T}$ and $\vec{W} = \sum_t \vec{W}_t \frac{\Delta t}{T}$ have dynamics
\begin{equation*}
\begin{split}
\frac{d}{d\tau} \vec{b} &= \eta \left\{  - \frac{\alpha}{T} \vec{b}    + g \vec{V}^{-1} ( \vec{b}^f - \vec{b}) \frac{e^{- \frac{1}{2} (\Delta \vec{b})^T \vec{V}^{-1} (\Delta \vec{b})}  }{\sqrt{(2 \pi)^D \det \vec{V}}} \right\} \\
\frac{d}{d\tau} \vec{W} &= \eta \left\{   - \frac{\alpha}{T}  \vec{W} \vec{\Sigma}   + g \left[    \vec{I} - \vec{V}^{-1} (\vec{b}^f - \vec{b}) (\vec{b}^f - \vec{b})^T  \right] \vec{V}^{-1} (\vec{W}^f - \vec{W}) \vec{\Sigma}  \ \frac{e^{- \frac{1}{2} (\Delta \vec{b})^T \vec{V}^{-1} (\Delta \vec{b})} }{\sqrt{(2 \pi)^D \det \vec{V}}}    \right\} \ .
\end{split}
\end{equation*}
As when we discussed pure bias learning above, it is conceptually helpful to define the \textit{hitting probability}
\begin{equation}
p_{\text{hit}}(\vec{b}, \vec{W}) := \frac{e^{- \frac{1}{2} (\Delta \vec{b})^T \vec{V}^{-1} (\Delta \vec{b})} }{\sqrt{(2 \pi)^D \det \vec{V}}}    \ ,
\end{equation}
whose size determines whether learning dynamics are fast or slow. In particular, the speed of learning dynamics depends on $(\Delta \vec{b})^T \vec{V}^{-1} (\Delta \vec{b})$: when the learned (endpoint) bias is sufficiently close to its target, learning is fast, and when it is not, learning can be extremely slow.

\clearpage

\section{Convergence time analysis for pure bias learning}
\label{app:conv}
As derived in Appendix \ref{app:dyn_deriv}, when there are no features, the endpoint-associated bias $\vec{b} = \sum_t \vec{b}_t \frac{\Delta t}{T}$ evolves according to
\begin{equation} 
\frac{d}{d\tau} \vec{b} = \eta \left\{  - \frac{\alpha}{T} \vec{b}   + g \tilde{\vec{B}} ( \vec{b}^f - \vec{b}) \ \sqrt{\frac{\det\tilde{\vec{B}}}{(2 \pi)^D}} e^{- \frac{1}{2} (\vec{b} - \vec{b}^f)^T \tilde{\vec{B}} (\vec{b} - \vec{b}^f)}  \right\} \ .
\end{equation}
Assuming the effort term is negligible, and that $\tilde{\vec{B}} = \tilde{\beta} \vec{I}$, we have
\begin{equation} 
\boxed{ \ \frac{d}{d\tau} \vec{b} \approx  \eta g \tilde{\beta} ( \vec{b}^f - \vec{b}) \ \left( \frac{\tilde{\beta}}{2 \pi} \right)^{D/2} \ e^{- \frac{\tilde{\beta}}{2} \Vert \vec{b} - \vec{b}^f \Vert_2^2}  \ } \ .
\end{equation}
In this appendix, we analytically characterize the convergence time of this ODE.

\subsection{Evolution of bias error}

The trick is to directly track the time evolution of the \textit{bias error}
\begin{equation}
E := \tilde{\beta} \Vert \vec{b} - \vec{b}^f \Vert_2^2 \ .
\end{equation}
Its time derivative is
\begin{equation}
\frac{d}{d\tau} E = 2 \tilde{\beta} (\vec{b} - \vec{b}^f)^T \frac{d \vec{b}}{d\tau} = - 2 \eta g \tilde{\beta}  \left(  \frac{\tilde{\beta}}{2 \pi} \right)^{D/2}  \left[  \tilde{\beta} \Vert \vec{b} - \vec{b}^f \Vert_2^2  \right]   \ e^{- \frac{\tilde{\beta}}{2} \Vert \vec{b} - \vec{b}^f \Vert_2^2}  \ ,
\end{equation}
so it evolves according to the ODE
\begin{equation}
\boxed{ \ \frac{d}{d\tau} E = - 2 \eta g \tilde{\beta} \left(  \frac{\tilde{\beta}}{2 \pi} \right)^{D/2} \ E e^{- E/2} \ } \ .
\end{equation}
We can solve this ODE by separation of variables. If $E$ evolves from an initial error $E_0$ at $\tau = 0$ to a final error $E_f$ at $\tau = t_*$, then
\begin{equation}
\begin{split}
\int_{E_0}^{E_f} \frac{e^{E/2}}{E} \ dE &= - 2 \eta g \tilde{\beta} \left(  \frac{\tilde{\beta}}{2 \pi} \right)^{D/2} \int_0^{t_*} d\tau \\
\implies \ \int_{E_0/2}^{E_f/2} \frac{e^{u}}{u} \ du &= - 2 \eta g \tilde{\beta} \left(  \frac{\tilde{\beta}}{2 \pi} \right)^{D/2} t_* \\
\implies \ \text{Ei}( E_f/2) - \text{Ei}( E_0/2) &= - 2 \eta g \tilde{\beta} \left(  \frac{\tilde{\beta}}{2 \pi} \right)^{D/2} t_* \ ,
\end{split}
\end{equation}
where $\text{Ei}(x) := \int_{-\infty}^x \frac{e^y}{y} dy$ is the \textit{exponential integral function}, a quickly growing special function. Rearranging, we find that
\begin{equation} \label{eq:conv_time_app}
\boxed{ \ t_* =  \frac{1}{2 \eta g \tilde{\beta}} \left( \frac{2 \pi}{\tilde{\beta}} \right)^{D/2} \left[  \ \text{Ei}\left( \frac{\tilde{\beta}}{2} \Vert \vec{b}^0 - \vec{b}^f \Vert_2^2  \right) - \text{Ei}\left(  \frac{1}{2} \epsilon^2   \right) \ \right] \ }  \ ,
\end{equation}
where we used $E_f = \epsilon^2$ to denote the final (small) bias error, and $\vec{b}^0$ to denote the initial endpoint bias.

\subsection{Initial bias error in terms of $D$ and $M$}

Recall that a major objective of our work is to characterize how the task's spatial dimensionality $D$ and number of time steps $M$ affect the speed and dynamics of learning. Since convergence time depends only on the initial error $\tilde{\beta} \Vert \vec{b}^0 - \vec{b}^f \Vert_2^2$ here, answering this question (at least in this setting) amounts to determining how this error scales with $D$ and $M$.

The relationship between $\vec{b}^0$ and $D$ and $M$ depends on how we choose to initialize our agent. We consider two typical desiderata here:
\begin{itemize}
\item Variance in the initial endpoint $\vec{x}_T := \sum_{t = 0}^{M-1} \vec{u}_t \Delta t$ must be $\mathcal{O}(1)$ with respect to $M$.
\item Variance in the initial endpoint should scale linearly with $D$.
\end{itemize}
The first condition ensures that the initial (average) distance between the agent and the target does not blow up at the number of time steps increases. The second condition  is natural because, if initializations along each direction are independent, linear increases in dimensionality ought to linearly increase the overall (squared) distance. 

To make things concrete, consider a typical Gaussian initialization for the bias $b_{t i}$ associated with the average action at time step $t \in \{ 0, ..., M-1 \}$ along direction $i \in \{ 1, ..., D \}$. If
\begin{equation}
b^0_{t i} \sim \mathcal{N}\left( 0, \xi^2 \frac{T}{\Delta t} \right) 
\end{equation}
for some scalar $\xi > 0$ controlling the overall variance in the initialization, then the covariance matrix associated with the endpoint bias $\vec{b}^0 = \sum_t \vec{b}^0_t \frac{\Delta t}{T}$ is
\begin{equation}
\mathbb{E}_{\vec{b}^0_{0:T}}[ \vec{b}^0 (\vec{b}^0)^T ] = \sum_{t, t'} \frac{(\Delta t)^2}{T^2} \mathbb{E}[ \vec{b}^0_t (\vec{b}^0_{t'})^T ] = \sum_{t, t'} \frac{(\Delta t)^2}{T^2} \delta_{t, t'} \xi^2 \frac{T}{\Delta t} \vec{I} = \xi^2 \vec{I} \ , 
\end{equation}
i.e., it is $\mathcal{O}(1)$ with respect to $M$. But the average norm of $\vec{b}^0$ depends linearly on $D$, since
\begin{equation}
\mathbb{E}_{\vec{b}^0_{0:T}}[ \Vert \vec{b}^0  \Vert_2^2  ] = \text{tr}[ \xi^2 \vec{I} ] = \xi^2 D \ .
\end{equation}
Incidentally, the $(T/\Delta t)$ scaling of the variance of $b^0_{t i}$ was necessary to include, since without it both the covariance and norm would scale like $1/M$. This would cause the agent to always start with parameters near zero in the large $M$ limit.

Analogously, assume the components of the target location $\vec{b}^f$ along each direction are chosen independently from a Gaussian. Concretely, assume
\begin{equation}
b^f_i \sim \mathcal{N}(0, \xi_f^2)
\end{equation}
for some $\xi_f > 0$. 

Given these Gaussian initializations, a typical value of the initial error is
\begin{equation}
\tilde{\beta} \ \mathbb{E}[ \Vert \vec{b}^0 - \vec{b}^f \Vert_2^2 ] = \tilde{\beta} \left\{ \mathbb{E}[ \Vert \vec{b}^0 \Vert_2^2 ] + \mathbb{E}[\Vert \vec{b}^f \Vert_2^2]     \right\} = \tilde{\beta} D \left\{ \xi^2 + \xi_f^2    \right\} \ .
\end{equation}
In summary, the initial error ought to scale linearly with $D$ but not at all with $M$. We can heuristically write typical values of the initial error as $\tilde{\beta} D d_0^2$, where $d_0$ is a `typical' (i.e., across spatial dimensions) initial distance between the agent's endpoint bias and the target bias.



\subsection{Dense reward regime} 

The convergence time (Eq. \ref{eq:conv_time_app}) behaves substantively differently depending on the size of the initial error $E_0$; we show this explicitly in this subsection and the next.

Suppose the initial error $\tilde{\beta} \Vert \vec{b}^0 - \vec{b}^f \Vert_2^2$ is small, so that agent consistently hits the target; because it receives rewards frequently when this condition holds, we call this the \textit{dense reward regime}.

The exponential integral has a small argument expansion
\begin{equation}
\text{Ei}(x) = \gamma + \log x + x + \frac{x^2}{4} + \cdots
\end{equation}
where $\gamma = 0.5772...$ is the Euler-Mascheroni constant. This implies that the convergence time is approximately
\begin{equation}
\begin{split}
t_* &\approx  \frac{1}{2 \eta g \tilde{\beta}} \left( \frac{2 \pi}{\tilde{\beta}} \right)^{D/2} \left[  \ \gamma + \log\left( \frac{\tilde{\beta}}{2} \Vert \vec{b}^0 - \vec{b}^f \Vert_2^2  \right) - \gamma - \log\left(  \frac{1}{2} \epsilon^2   \right) \ \right] \\
&=   \frac{1}{2 \eta g \tilde{\beta}} \left( \frac{2 \pi}{\tilde{\beta}} \right)^{D/2}  \log\left( \frac{\tilde{\beta} \Vert \vec{b}^0 - \vec{b}^f \Vert_2^2}{\epsilon^2}  \right)  \  .
\end{split}
\end{equation}
Assuming $E_0 \sim \tilde{\beta} D d_0^2$, this becomes
\begin{equation}
\boxed{ \ t_* \approx  \frac{1}{2 \eta g \tilde{\beta}} \left( \frac{2 \pi}{\tilde{\beta}} \right)^{D/2}  \log\left( \frac{\tilde{\beta} D d_0^2}{\epsilon^2}  \right) \ }  \ ,
\end{equation}
i.e., we find that the convergence time scales \textit{logarithmically} with the spatial dimensionality $D$ and squared distance $d_0^2$. This means that large increases in either quantity do not produce large increases in convergence time; in this sense, learning is relatively easy and fast.

\subsection{Sparse reward regime}

Suppose the initial error $\tilde{\beta} \Vert \vec{b}^0 - \vec{b}^f \Vert_2^2$ is large, so that the agent initially only rarely hits the target. We call this the \textit{sparse reward regime}.

The exponential integral has a large argument expansion
\begin{equation}
\text{Ei}(x) = \frac{e^x}{x} \left( 1 + \frac{1}{x} + \frac{2}{x^2} + \cdots \right) \ .
\end{equation}
This implies that the convergence time is approximately
\begin{equation} \label{eq:conv_slow_app}
 t_* \approx  \frac{1}{\eta g \tilde{\beta}^2} \left( \frac{2 \pi}{\tilde{\beta}} \right)^{D/2}   \ \frac{e^{ \frac{\tilde{\beta}}{2} \Vert \vec{b}^0 - \vec{b}^f \Vert_2^2}}{\Vert \vec{b}^0 - \vec{b}^f \Vert_2^2}  \ ,
\end{equation}
where we have neglected the (much smaller) $\epsilon$-dependent term. Assuming $E_0 \sim \tilde{\beta} D d_0^2$, this becomes
\begin{equation} 
\boxed{ \ t_* \approx  \frac{1}{\eta g \tilde{\beta}^2} \left( \frac{2 \pi}{\tilde{\beta}} \right)^{D/2}   \ \frac{e^{ \frac{\tilde{\beta}}{2} D d_0^2}}{D d_0^2} \ } \ ,
\end{equation}
which means that the convergence time scales \textit{exponentially} with the spatial dimensionality $D$ and the squared distance $d_0^2$. This means that small increases in either can produce large increases in learning time; in this sense, learning is extremely slow in general.


\clearpage

\section{Trajectory refinement after learning to reach target}
\label{app:temp}

In this appendix, we characterize the refinement of trajectories after agents learn to reliably hit the target. As derived in Appendix \ref{app:dyn_deriv}, when there are no features, the biases $\vec{b}_t$ evolve according to
\begin{equation} 
\frac{d}{d\tau} \vec{b}_t = \eta \left\{  - \frac{\alpha}{T} \vec{b}_t   + g \tilde{\vec{B}} ( \vec{b}^f - \vec{b}) \ \sqrt{\frac{\det\tilde{\vec{B}}}{(2 \pi)^D}} e^{- \frac{1}{2} (\vec{b} - \vec{b}^f)^T \tilde{\vec{B}} (\vec{b} - \vec{b}^f)}  \right\} \ .
\end{equation}
When $\vec{b}$ is \textit{sufficiently close} to the target bias $\vec{b}^f$, it is true that
\begin{equation}
e^{- \frac{1}{2} (\vec{b} - \vec{b}^f)^T \tilde{\vec{B}} (\vec{b} - \vec{b}^f)} \approx 1 \ ,
\end{equation}
so we obtain a \textit{linear} ODE
\begin{equation} \label{eq:bias_linear_app}
\boxed{ \ \frac{d}{d\tau} \vec{b}_t = \eta \left\{  - \frac{\alpha}{T} \vec{b}_t   + g \tilde{\vec{B}} ( \vec{b}^f - \vec{b}) \ \sqrt{\frac{\det\tilde{\vec{B}}}{(2 \pi)^D}}  \right\} \ } \ .
\end{equation}
We will characterize the trajectory refinement phase by solving this ODE explicitly.

\subsection{Decoupling space- and time-dependence}

Eq. \ref{eq:bias_linear_app} generically couples $b_{t i}$ (the bias parameter associated with the time step $t$ mean action along direction $i$) to $b_{t' j}$ for various $t' \neq t$ and $j \neq i$. We can greatly simplify dynamics by rotating them so that they become less coupled to one another.

Let $\tilde{\vec{B}} = \vec{O} \vec{\Lambda}_B \vec{O}^T$ be an eigendecomposition of $\tilde{\vec{B}}$, so that $\vec{O}$ is orthogonal (i.e., $\vec{O} \vec{O}^T = \vec{O}^T \vec{O} = \vec{I}$) and $\vec{\Lambda}_B$ is diagonal. Let $\tilde{\beta}_i = \Lambda_{B, ii}$ denote the $i$-th eigenvalue of $\tilde{\vec{B}}$. If we define rotated biases $\tilde{\vec{b}}_t := \vec{O}^T \vec{b}_t$, and the corresponding rotated endpoint $\tilde{\vec{b}} := \vec{O}^T \vec{b}$ and target $\tilde{\vec{b}}^f := \vec{O}^T \vec{b}^f$, then
\begin{equation}
\frac{d}{d\tau} \tilde{\vec{b}}_t = \eta \left\{  - \frac{\alpha}{T} \tilde{\vec{b}}_t   + g \ \vec{O}^T \tilde{\vec{B}} \vec{O}  ( \tilde{\vec{b}}^f - \tilde{\vec{b}}) \ \sqrt{\frac{\det\tilde{\vec{B}}}{(2 \pi)^D}}  \right\} =  \eta \left\{  - \frac{\alpha}{T} \tilde{\vec{b}}_t   + g \ \vec{\Lambda}_B ( \tilde{\vec{b}}^f - \tilde{\vec{b}}) \ \sqrt{\frac{\det\tilde{\vec{B}}}{(2 \pi)^D}}  \right\}  \ .
\end{equation}
Since $\vec{\Lambda}_B$ is diagonal, the spatial components of this ODE decouple, i.e.,
\begin{equation}
\frac{d}{d\tau} \tilde{b}_{t i} = \eta \left\{  - \frac{\alpha}{T} \tilde{b}_{t i}   + g \tilde{\beta}_i ( \tilde{b}^f_i - \tilde{b}_i) \ \sqrt{\frac{\det\tilde{\vec{B}}}{(2 \pi)^D}}  \right\}  \ .
\end{equation}
To ease notation, define the effective gain 
\begin{equation}
g_i := g \tilde{\beta}_i \sqrt{\frac{\det\tilde{\vec{B}}}{(2 \pi)^D}}
\end{equation}
and leave the spatial index on all quantities implicit. We then have
\begin{equation} \label{eq:matrix_baby}
\boxed{ \ \frac{d}{d\tau} \tilde{b}_{t} = \eta \left\{  - \frac{\alpha}{T} \tilde{b}_{t}   + g_i \left( \tilde{b}^f - \sum_{t = 0}^{M-1} \tilde{b}_t \frac{\Delta t}{T} \right) \  \right\}  \ } \ .
\end{equation}

\subsection{Solving the ODE} 

A nice feature of Eq. \ref{eq:matrix_baby} is that it linearly couples the $M$ quantities $\tilde{b}_0, ..., \tilde{b}_{M-1}$, and hence can be recast as a matrix ODE. Define the vector $\vec{y} := (\tilde{b}_0, ..., \tilde{b}_{M-1})^T \in \mathbb{R}^M$ containing the biases, and define the matrix
\begin{equation}
\vec{A} := \frac{\alpha}{T} \vec{I} + g_i \frac{\Delta t}{T} \vec{1} \vec{1}^T \ .
\end{equation}
Eq. \ref{eq:matrix_baby} is equivalent to the matrix ODE
\begin{equation} \label{eq:matrix_ODE}
\frac{d}{d\tau} \vec{y} = \eta  ( g_i \tilde{b}^f \vec{1} - \vec{A} \vec{y}) = \eta \vec{A} (\vec{y}^* - \vec{y}) \ ,
\end{equation}
where we use $\vec{y}^* := g_i \tilde{b}^f \vec{A}^{-1} \vec{1}$ to denote the fixed point.

\paragraph{Fixed point.} Since the inverse of $\vec{A}$ is
\begin{equation}
\vec{A}^{-1} = \frac{1}{\alpha/T} \vec{I} - \frac{\frac{g_i \Delta t}{T}}{(\alpha/T)^2} \frac{\vec{1} \vec{1}^T}{1 + \frac{1}{\alpha/T} \frac{g_i \Delta t}{T} M} = \frac{1}{\alpha/T} \left(  \vec{I} - \frac{\frac{g \Delta t}{T}}{\alpha/T + g_i} \vec{1} \vec{1}^T \right)  \ ,
\end{equation}
the fixed point $\vec{y}^*$ occurs at
\begin{equation}
\vec{y}^* := g_i \tilde{b}^f \vec{A}^{-1} \vec{1} = \frac{g_i \tilde{b}^f}{\alpha/T} \left(  1 - \frac{g_i}{\alpha/T + g_i} \right) \vec{1} = \frac{g_i}{\alpha/T + g_i} \tilde{b}^f \ \vec{1} \ .
\end{equation}
This gets increasingly close to $\tilde{b}^f$ as the coefficient $\alpha/T$ of the effort term gets smaller.

\paragraph{Eigenvalues and eigenvectors.} The characteristic polynomial of $\vec{A}$ is
\begin{equation}
\begin{split}
\det( \vec{A} - \lambda \vec{I} ) &= \det\left[ \left( \frac{\alpha}{T} - \lambda \right) \vec{I} + \frac{g_i \Delta t}{T} \vec{1} \vec{1}^T \right] \\
&= \left(  \frac{\alpha}{T} - \lambda \right)^M \left(  1 + \frac{g_i \Delta t}{T} \frac{M}{ \frac{\alpha}{T} - \lambda}    \right) \\
&= \left(  \frac{\alpha}{T} - \lambda \right)^{M-1} \left(   \frac{\alpha}{T} - \lambda + g_i  \right) \ ,
\end{split}
\end{equation}
so $\vec{A}$ has two distinct eigenvalues: a `bulk' eigenvalue $\lambda_{\text{bulk}} := \alpha/T$ (with multiplicity $M-1$), and an `outlier' eigenvalue $\lambda_* := \alpha/T + g_i$ (with multiplicity $1$). The fact that there is a gap between these eigenvalues, which is determined by the size of $g_i > 0$, implies that the component of the solution along the eigenvector associated with $\lambda_*$ is learned before other components. 

Since $\vec{A}$ has a particularly simple structure---it is just a scaled identity matrix plus a rank-one perturbation---we can determine the corresponding eigenvectors by inspection. $\vec{1}$ is an eigenvector, since
\begin{equation}
\vec{A} \vec{1} = \left(  \frac{\alpha}{T} \vec{I} + \frac{g_i \Delta t}{T} \vec{1} \vec{1}^T \right) \vec{1} = \frac{\alpha}{T} \vec{1} + \frac{g_i \Delta t M}{T} \vec{1} = \left( \frac{\alpha}{T} + g_i \right) \vec{1} \ . 
\end{equation}
Moreover, $\vec{1}$ is the outlier eigenvector which is learned first, since its corresponding eigenvalue is the outlier $\lambda_* = \alpha/T + g_i$. This is significant because the endpoint is proportional to $\vec{1}$:
\begin{equation}
\tilde{b} = \sum_t \tilde{b}_t \frac{\Delta t}{T} = \frac{\Delta t}{T} \vec{y}^T \vec{1} \ . 
\end{equation}
Hence, the endpoint is learned before other trajectory components, with the difference in learning time determined by the eigenvalue gap.

The other eigenvectors are simple: any vector orthogonal to $\vec{1}$ is an eigenvector with eigenvalue $\lambda_{\text{bulk}} = \alpha/T$.

\paragraph{Solution.} Our matrix ODE (Eq. \ref{eq:matrix_ODE}) has the formal solution
\begin{equation}
\vec{y}(\tau) = e^{- \eta \vec{A} \tau} \vec{y}(0) + (\vec{I} - e^{- \eta \vec{A} \tau}) \vec{y}^* = \vec{y}^* + e^{- \eta \vec{A} \tau} [ \vec{y}(0) - \vec{y}^* ] 
\end{equation}
in terms of the matrix exponential $e^{-\eta \vec{A} \tau}$. But the simple structure of $\vec{A}$ allows us to write it more explicitly. The matrix exponential that appears can be rewritten as
\begin{equation}
e^{- \eta \vec{A} \tau}  = e^{- \frac{\eta \tau \alpha}{T}} e^{- \eta \tau g_i \frac{\Delta t}{T} \vec{1} \vec{1}^T} = e^{- \frac{\eta \tau \alpha}{T}} \left[  \vec{I} + \frac{\Delta t}{T} ( e^{- \eta \tau g_i } - 1 ) \vec{1} \vec{1}^T \right] \ .
\end{equation}
Let $\tilde{b} = \frac{\Delta t}{T} \vec{1}^T \vec{y}$ denote the (rotated) learned endpoint, and $\tilde{b}_{\perp}$ any component of the solution orthogonal to $\vec{1}$. We conclude that these evolve according to
\begin{equation}
\begin{split}
\tilde{b}(\tau) &= e^{- \eta ( \alpha/T + g_i ) \tau}  \ \tilde{b}(0) + ( 1 - e^{- \eta ( \alpha/T + g_i ) \tau} ) \frac{g_i}{\alpha/T + g_i} \tilde{b}^f \\
\tilde{b}_{\perp}(\tau) &= e^{- \eta (\alpha/T ) \tau} \  \tilde{b}_{\perp}(0) \ .
\end{split}
\end{equation}
In words: the non-endpoint components converge to their target value (i.e., zero) on a slower time scale $\sim ( \eta \lambda_{\text{bulk}})^{-1} = T/(\eta \alpha)$. 

%

\subsection{Total vs endpoint error}

Another way to quantify the trajectory refinement effect is to track how bias errors evolve in time. We briefly do this below.

\paragraph{Optimal trajectory.} Restoring spatial indices for clarity, our fixed point result above says that the optimal $\tilde{b}_{t i}$ equals
\begin{equation}
\tilde{b}^*_{t i} = \frac{g_i}{\alpha/T + g_i} \tilde{b}^f_i \ .
\end{equation}
Define the diagonal matrix $\vec{D}$ whose diagonal entries are
\begin{equation}
D_{ii} := \frac{g_i}{\alpha/T + g_i} \ .
\end{equation}
In (spatial) vector form, this result reads
\begin{equation}
\tilde{\vec{b}}^*_t = \vec{D} \tilde{\vec{b}}^f \ ,
\end{equation}
which implies that the non-rotated optimal trajectory has
\begin{equation}
\vec{b}_t^* = \vec{O} \vec{D} \vec{O}^T \vec{b}^f \ .
\end{equation}
Importantly, \textit{it does not depend on time}: the trajectory which optimizes the return uses the same velocity input at all time steps, since this choice reaches the target while minimizing the effort term. 

\paragraph{Trajectory error decomposition.} Since we know that the bias at time step $t$ ought to equal $\vec{b}^*_t$, we can define the \textit{total trajectory error}
\begin{equation}
E_{\text{tot}} := \sum_{t = 0}^{M-1} (\vec{b}_t - \vec{b}_t^*)^T \tilde{\vec{B}}  (\vec{b}_t - \vec{b}_t^*) \ .
\end{equation}
Reusing the eigendecomposition $\tilde{\vec{B}} = \vec{O} \vec{\Lambda}_B \vec{O}^T$, we can rewrite the error as
\begin{equation}
E_{\text{tot}} = \sum_{i = 1}^D \tilde{\beta}_i \sum_{t = 0}^{M-1} ( \tilde{b}_{t i} - D_i \tilde{b}^f_i )^2 \ .
\end{equation}
Let $\tilde{b}_i := \sum_t \tilde{b}_{t i} \frac{\Delta t}{T}$ denote the (rotated) endpoint bias along (rotated) direction $i$. We immediately have a bias-variance-type decomposition
\begin{equation}
E_{\text{tot}} = \sum_{i = 1}^D \tilde{\beta}_i \sum_{t = 0}^{M-1} ( \tilde{b}_{t i} - \tilde{b}_i + \tilde{b}_i - D_i \tilde{b}^f_i )^2 =  E_{\text{end}} + E_{\text{var}}
\end{equation}
where $E_{\text{end}}$ is the \textit{endpoint error} (the `bias' term) and $E_{\text{var}}$ is the \textit{trajectory variance} (the `variance' term), with
\begin{equation}
\begin{split}
E_{\text{end}} &:= \sum_{i = 1}^D \tilde{\beta}_i ( \tilde{b}_{i} - D_{ii} \tilde{b}^f_i  )^2 = \sum_{i = 1}^D \tilde{\beta}_i \left[ \frac{\Delta t}{T} (\vec{1}^T \vec{y}_i) - D_{ii} \tilde{b}_i^f \right]^2  \\
E_{\text{var}} &:=  \sum_{i = 1}^D \tilde{\beta}_i \sum_{t = 0}^{M-1} \frac{\Delta t}{T} ( \tilde{b}_{t i} - \tilde{b}_i  )^2 = \sum_{i = 1}^D \tilde{\beta}_i  \Vert \vec{y}_i - \frac{\Delta t}{T} (\vec{1}^T \vec{y}_i) \vec{1} \Vert_2^2 \ .
\end{split}
\end{equation}

\paragraph{Time scales of error evolution.} Above, we found that (at least in the `dense' reward regime) there are two time scales of trajectory learning: learning the \textit{endpoint}, which occurs at a rate $\eta \lambda_*$; and learning all other components, which occurs at a rate $\eta \lambda_{\text{bulk}} \ll \eta \lambda_*$. 

We can make this difference manifestly clear if we write $\vec{y}_i$ in terms of eigenvectors of $\vec{A}_i$, i.e.,
\begin{equation}
\vec{y}_i(\tau) = c^i_0(\tau) \vec{1} + \sum_{k = 1}^{M-1} c^i_k(\tau) \vec{v}_k
\end{equation}
where the $\vec{v}_k$ are normalized vectors orthogonal to each other and orthogonal to $\vec{1}$. (Note that we can use the same eigenvectors for all $i$, since all $\vec{A}_i$ have the same eigenvectors.) Our matrix ODE (Eq. \ref{eq:matrix_ODE}) implies
\begin{equation}
\begin{split}
\frac{d}{d\tau} c^i_0 &=  \eta \lambda_* ( D_{ii} \tilde{b}^f_i  - c^i_0) \\
\frac{d}{d\tau} c^i_k &= - \eta \lambda_{\text{bulk}} c^i_k \ .
\end{split}
\end{equation}
Then since $\frac{\Delta t}{T} (\vec{1}^T \vec{y}_i) = c^i_0$, the variance term equals
\begin{equation}
E_{\text{var}} = \sum_{i = 1}^D \tilde{\beta}_i \Vert \sum_{k = 1}^{M-1} c^i_k(\tau) \vec{v}_k \Vert_2^2 = \sum_{i = 1}^D \tilde{\beta}_i \sum_{k = 1}^{M-1} [c_k^i(\tau)]^2 
\end{equation}
and has
\begin{equation}
\frac{d}{d\tau} E_{\text{var}} = - \sum_{i = 1}^D \tilde{\beta}_i 2 \eta \lambda_{\text{bulk}} \sum_{k = 1}^{M-1} [c_k^i(\tau)]^2 = - 2 \eta \lambda_{\text{bulk}} E_{\text{var}} \ .
\end{equation}
Meanwhile, the bias term equals
\begin{equation}
E_{\text{end}} = \sum_{i = 1}^D \tilde{\beta}_i \left[ c^i_0(\tau) - D_{ii} \tilde{b}^f_i \right]^2
\end{equation}
and has
\begin{equation}
\frac{d}{d\tau} E_{\text{end}} = - 2 \eta \lambda_* \sum_{i = 1}^D \tilde{\beta}_i \left[ c^i_0(\tau) - D_{ii} \tilde{b}^f_i \right]^2 = - 2 \eta \lambda_* E_{\text{end}} \ .
\end{equation}
In summary, the total trajectory error decomposes as a sum of the \textit{endpoint error} and \textit{trajectory variance}, and these evolve at different rates:
\begin{equation}
\boxed{ \ \frac{d}{d\tau} E_{\text{end}} = - 2 \eta \lambda_{*} E_{\text{end}}  \hspace{0.5in}    \frac{d}{d\tau} E_{\text{var}} = - 2 \eta \lambda_{\text{bulk}} E_{\text{var}}  \ } \ .
\end{equation}
Moreover, the endpoint error converges the fastest, and the discrepancy between the two rates grows as the coefficient $\alpha/T$ of the effort term gets smaller.

\clearpage

\section{Optimal curricula for pure bias learning}
\label{app:optimal}

In this appendix, we derive optimal curricula for pure bias learning. If the effort term is negligibly small, the endpoint $\vec{b} = \sum_t \vec{b}_t \frac{\Delta t}{T}$ has dynamics
\begin{equation}  \label{eq:b_curr}
\boxed{ \ \frac{d}{d\tau} \vec{b} = \eta  g \tilde{\vec{B}} ( \vec{b}^f - \vec{b}) \ \sqrt{\frac{\det\tilde{\vec{B}}}{(2 \pi)^D}} e^{- \frac{1}{2} (\vec{b} - \vec{b}^f)^T \tilde{\vec{B}} (\vec{b} - \vec{b}^f)} \ } 
\end{equation}
where the $D \times D$ positive definite matrix
\begin{equation}
\tilde{\vec{B}}^{-1} := \frac{1}{\kappa} \vec{I} + \sum_t \vec{B}_t^{-1} \Delta t 
\end{equation}
controls the \textit{effective size} of the target, or equivalently the size of the hitting probability factor. Intuitively, we would like to choose the matrix $\tilde{\vec{B}}$ so that the hitting probability consistently remains large.

Define the $D \times D$ positive semidefinite matrix $\vec{B}$ via
\begin{equation}
\vec{B}^{-1} := \sum_t \vec{B}_t^{-1} \Delta t  \ .
\end{equation}
The parameter $\kappa$ (which determines the `intrinsic size' of the target) cannot be controlled, but we will assume that the exploration matrices $\vec{B}_t$ can be. Since they only affect dynamics through $\vec{B}$, we will assume that $\vec{B}$ is controllable. How should we choose $\vec{B}$ in order to reduce the convergence time as much as possible? Equivalently, what is the optimal \textit{curriculum} for $\vec{B}$?

%
%


In order to formalize `as much as possible' here, we exploit the fact that dynamics correspond to gradient ascent up an effective return (see Appendix \ref{app:dyn_deriv}). Ignoring the effort term, the effective return is
\begin{equation}
G_{\text{eff}}(\vec{b}, \vec{B}) := g \sqrt{\frac{\det \tilde{\vec{B}}}{(2 \pi)^D}} \ e^{- \frac{1}{2} \left(  \vec{b} - \vec{b}^f \right)^T \tilde{\vec{B}} \left(  \vec{b} - \vec{b}^f \right) }  
\end{equation}
and depends on policy parameters entirely through $\vec{b}$. Note that Eq. \ref{eq:b_curr} can also be written as
\begin{equation}
\frac{d}{d\tau} \vec{b} = \eta \frac{\partial G_{\text{eff}}}{\partial \vec{b}}  \ .
\end{equation}
Since the dynamics of $\vec{b}$ are `greedy' with respect to this function, in the sense that they move $\vec{b}_t$ in the direction that increases it the most, we can define an `optimal' $\vec{B}$ as one which also maximizes $G_{\text{eff}}$.


\subsection{One-dimensional case}

For simplicity, we will begin with the $D = 1$ case, which has a scalar exploration parameter
\begin{equation}
\frac{1}{\tilde{\beta}} := \frac{1}{\kappa} + \sum_{t=0}^{M-1} \frac{1}{\beta_t} \Delta t = \frac{1}{\kappa} + \frac{1}{\beta} \ .
\end{equation}
In one dimension, the effective return is
\begin{equation}
G_{\text{eff}}(b, \beta) = g \sqrt{\frac{\tilde{\beta}}{2 \pi}} e^{- \frac{\tilde{\beta}}{2} ( b - b^f )^2}  \ ,
\end{equation}
and its derivative with respect to $\tilde{\beta}$ equals
\begin{equation}
\frac{\partial G_{\text{eff}}}{\partial \tilde{\beta}} = \frac{g}{\sqrt{2 \pi}} \frac{\partial}{\partial \tilde{\beta}}\left[  \tilde{\beta}^{1/2} e^{- \frac{\tilde{\beta}}{2} (b - b^f)^2}  \right] = \frac{g}{2 \sqrt{2 \pi}} \left[ \frac{1}{\tilde{\beta}} - (b - b^f)^2  \right]   \tilde{\beta}^{1/2} e^{- \frac{\tilde{\beta}}{2} (b - b^f)^2} \ .
\end{equation}
Its derivative with respect to $\beta$ equals
\begin{equation}
\frac{\partial G_{\text{eff}}}{\partial \beta} = \frac{\partial G_{\text{eff}}}{\partial \tilde{\beta}} \frac{\partial \tilde{\beta}}{\partial \beta}  \ ,
\end{equation}
but since $\tilde{\beta}$ is a monotonic function of $\beta$, these two derivatives have the same critical points. They both equal zero where
\begin{equation}
\boxed{ \ \frac{1}{\tilde{\beta}} = (b - b^f)^2 \ } \ .
\end{equation}
We call this a \textit{variance-matching condition}, since when it holds the argument of the hitting probability is order one. 

The optimal choice of $\beta$ does not \textit{always} satisfy this condition, however. Since it is equivalent to
\begin{equation}
\frac{1}{\kappa} + \frac{1}{\beta} = (b - b^f)^2  \ ,
\end{equation} 
it is impossible to satisfy if $\kappa (b - b^f)^2 < 1$. That is, it cannot be satisfied when $b$ is close enough to its optimal value that exploration cannot improve the hitting probability. In this case, the solution is instead to have no exploration, i.e., $\beta = \infty$. 

Finally, the optimal $\beta$ curriculum in the one-dimensional case is
\begin{equation} \label{eq:beta_1D_opt}
\boxed{ \ \beta_*( \Delta b ) = \begin{cases}
\frac{1}{(b - b^f)^2 - \frac{1}{\kappa}} & \text{ if } \kappa (b - b^f)^2 > 1 \\
\infty & \text{ if }  \kappa (b - b^f)^2 \leq 1 
\end{cases} \ } \ .
\end{equation}
Provided the bias error is sufficiently big, i.e., $\kappa (b - b^f)^2 > 1$, using this curriculum yields 
\begin{equation}
\tilde{\beta} ( b - b^f)^2 = 1 \ ,
\end{equation}
which implies that the hitting probability becomes
\begin{equation} \label{eq:hit_1D}
\boxed{ \ p_{\text{hit}} = \sqrt{\frac{1}{2 \pi (b - b^f)^2}} e^{-1/2} \ } \ .
\end{equation}

\subsection{$D$-dimensional case}

We will now study the higher-dimensional case---assume $D > 1$. First, to make the optimization slightly more tractable, we will maximize the logarithm of $G_{\text{eff}}$, i.e.,
\begin{equation}
\log G_{\text{eff}} = - \frac{1}{2} \log\det\left[ \frac{1}{\kappa} \vec{I} + \vec{B}^{-1} \right] - \frac{1}{2} (\Delta \vec{b})^T \left( \frac{1}{\kappa} \vec{I} + \vec{B}^{-1} \right)^{-1} (\Delta \vec{b}) + \text{const.} 
\end{equation}
\paragraph{Bias error is an eigenvector of the optimal matrix.} Second, a change of coordinates will make a subtle conceptual point clearer. Choose a coordinate system where one direction is aligned with $(\Delta \vec{b})$, and all other directions are orthogonal to it. If we let $\vec{e}$ denote the corresponding unit vector and $r := \Vert \Delta \vec{b} \Vert_2$, then $\Delta \vec{b} = r \vec{e}$. Now write $\tilde{\vec{B}}$ as a block matrix
\begin{equation}
\tilde{\vec{B}} = \begin{pmatrix}
a & \vec{c}^T \\
\vec{c} & \vec{G}
\end{pmatrix}
\end{equation}
where $a$ is a scalar, $\vec{c} \in \mathbb{R}^{D-1}$, and $\vec{G} \in \mathbb{R}^{(D-1)\times(D-1)}$. The determinant of $\tilde{\vec{B}}$ equals
\begin{equation}
\det \tilde{\vec{B}} = \det\vec{G} \left[ a - \vec{c}^T \vec{G}^{-1} \vec{c} \right] \ .
\end{equation}
This means that, in this coordinate system (and in terms of $\tilde{\vec{B}}$ rather than $\vec{B}$ for simplicity), $\log G_{\text{eff}}$ equals
\begin{equation} \label{eq:lg_inter}
\log G_{\text{eff}} = \frac{1}{2} \log\det\vec{G} + \frac{1}{2} \log\left[ a - \vec{c}^T \vec{G}^{-1} \vec{c} \right] - \frac{r^2}{2} a + \text{const.}
\end{equation}
Since $\vec{G}$ and its inverse are positive definite,
\begin{equation}
\frac{1}{2} \log\left[ a - \vec{c}^T \vec{G}^{-1} \vec{c} \right] < \frac{1}{2} \log a
\end{equation}
whenever $\vec{c} \neq \vec{0}$. This implies that we can maximize the effective return by choosing $\vec{c} = \vec{0}$, i.e., by not mixing components associated with the bias error $\Delta \vec{b}$ and those associated with orthogonal directions. Put differently, it is optimal for $\Delta \vec{b}$ to be an eigenvector of $\tilde{\vec{B}}$, and hence of $\vec{B}$. 

\paragraph{$\vec{B}^{-1}$ should be low-rank.} Eq. \ref{eq:lg_inter} indicates that we can maximize our objective by choosing the determinant of $\vec{G}$, the matrix describing $\tilde{\vec{B}}$ along all directions orthogonal to $\Delta \vec{b}$, to be as large as possible. But since
\begin{equation}
\tilde{\vec{B}}^{-1} = \frac{1}{\kappa} \vec{I} + \vec{B}^{-1} \ ,
\end{equation}
the eigenvalues of $\tilde{\vec{B}}^{-1}$ are those of $\vec{B}^{-1}$. If $\lambda$ is an eigenvalue of $\vec{B}^{-1}$, then $\lambda + 1/\kappa$ is the corresponding eigenvalue of $\tilde{\vec{B}}^{-1}$. Hence, we can explicitly compute that the log-determinant of $\tilde{\vec{B}}$ equals
\begin{equation}
\log\det\tilde{\vec{B}} = \sum_i \log\left(  \frac{1}{\lambda_i + 1/\kappa} \right) \ .
\end{equation}
The log-determinant of $\vec{G}$ has this form, except that a single eigenvalue (associated with $\Delta \vec{b}$) is left out. In any case, it is clear that we can maximize it by taking the $\lambda_i$ to be as small as possible, with
\begin{equation}
\sum_i \log\left(  \frac{1}{\lambda_i + 1/\kappa} \right) < \sum_i \log\left(  \kappa \right) 
\end{equation}
for nonzero $\lambda_i$. This implies that it is optimal for $\vec{B}^{-1}$ to be low-rank, and in particular to equal
\begin{equation} \label{eq:B_inv_low}
\vec{B}^{-1} = q (\Delta \vec{b}) (\Delta \vec{b})^T
\end{equation}
for some $q \geq 0$. 

\paragraph{Finishing the optimization.} We now only need to choose the $q$ coefficient that appears in Eq. \ref{eq:B_inv_low}. For this choice of $\vec{B}^{-1}$, we have
\begin{equation}
\tilde{\vec{B}}^{-1} =  \frac{1}{\kappa} \vec{I} + q (\Delta \vec{b}) (\Delta \vec{b})^T 
\end{equation}
and
\begin{equation}
\tilde{\vec{B}} = \left( \frac{1}{\kappa} \vec{I} + q (\Delta \vec{b}) (\Delta \vec{b})^T \right)^{-1} = \kappa \left[ \vec{I} - \frac{q \kappa}{1 + q \kappa \Vert \Delta \vec{b} \Vert_2^2} (\Delta \vec{b}) (\Delta \vec{b})^T  \right] \ .
\end{equation}
Evaluating the relevant determinant and quadratic form, the objective reads
\begin{equation}
\log G_{\text{eff}}(q) = - \frac{1}{2} \log\left[ 1 + \kappa q \Vert \Delta \vec{b} \Vert_2^2  \right] - \frac{1}{2} \frac{\kappa  \Vert \Delta \vec{b} \Vert_2^2}{1 + \kappa q \Vert \Delta \vec{b} \Vert_2^2} + \text{const.}
\end{equation}
Differentiating with respect to $q$, we find
\begin{equation}
\begin{split}
\frac{\partial \log G_{\text{eff}}}{\partial q} &= - \frac{1}{2} \frac{\kappa \Vert \Delta \vec{b} \Vert_2^2}{1 + \kappa q \Vert \Delta \vec{b} \Vert_2^2 } + \frac{1}{2} \kappa^2 \Vert \Delta \vec{b} \Vert_2^4 \frac{1}{( 1 + \kappa q \Vert \Delta \vec{b} \Vert_2^2 )^{2}} \\
&= - \frac{1}{2} \frac{\kappa \Vert \Delta \vec{b} \Vert_2^2}{1 + \kappa q \Vert \Delta \vec{b} \Vert_2^2 } \left[ 1 -\frac{\kappa \Vert \Delta \vec{b} \Vert_2^2}{1 + \kappa q \Vert \Delta \vec{b} \Vert_2^2 } \right] \\
&= - \frac{1}{2} \frac{\kappa \Vert \Delta \vec{b} \Vert_2^2}{1 + \kappa q \Vert \Delta \vec{b} \Vert_2^2 } \left[ \frac{1 + \kappa (q - 1) \Vert \Delta \vec{b} \Vert_2^2}{1 + \kappa q \Vert \Delta \vec{b} \Vert_2^2 } \right] \ .
\end{split}
\end{equation}
This equals zero where
\begin{equation}
q = 1 - \frac{1}{\kappa \Vert \Delta \vec{b} \Vert_2^2} \ .
\end{equation}
However, as in the one-dimensional case, it is not always possible to satisfy this condition. $\vec{B}^{-1}$ must be positive semidefinite, so we require that $q \geq 0$; the condition can only be satisfied if
\begin{equation} \label{eq:q_req}
\kappa \Vert \Delta \vec{b} \Vert_2^2 \geq 1 \  ,
\end{equation}
i.e., if the agent's bias is far enough from the target bias that increasing exploration can improve the hitting probability.

If it is possible to satisfy this condition, the optimal curriculum uses
\begin{equation}
\vec{B}^{-1} = \left( 1 - \frac{1}{\kappa \Vert \Delta \vec{b} \Vert_2^2} \right) (\Delta \vec{b}) (\Delta \vec{b})^T \ ,
\end{equation}
or equivalently the optimal $\tilde{\vec{B}}$ has
\begin{equation}
\begin{split}
\tilde{\vec{B}}^{-1} &= \frac{1}{\kappa} \vec{I} + \left( 1 - \frac{1}{\kappa \Vert \Delta \vec{b} \Vert_2^2} \right) (\Delta \vec{b}) (\Delta \vec{b})^T \\
\tilde{\vec{B}} &= \kappa \vec{I} - \left( \kappa - \frac{1}{\Vert \Delta \vec{b} \Vert_2^2} \right) \frac{(\Delta \vec{b}) (\Delta \vec{b})^T}{\Vert \Delta \vec{b} \Vert_2^2} \ .
\end{split}
\end{equation}
Additionally, the \textit{variance-matching condition}
\begin{equation}
\boxed{ \ (\Delta \vec{b})^T \tilde{\vec{B}} (\Delta \vec{b}) = 1 \ }
\end{equation}
holds.

If it is \textit{not} possible to satisfy Eq. \ref{eq:q_req}, then the above derivative with respect to $q$ indicates that $\log G_{\text{eff}}$ is a decreasing function of $q$. Hence, it is optimal to choose $q = 0$---i.e., to choose no exploration when the agent's policy parameters are close enough that exploration will not help.

Finally, the optimal exploration matrix $\vec{B}^{-1}$ equals
\begin{equation} \label{eq:B_opt_Ddim}
\boxed{ \ \vec{B}^{-1}_*(\Delta \vec{b}) = \begin{cases}
\left( 1 - \frac{1}{\kappa \Vert \Delta \vec{b} \Vert_2^2}\right) (\Delta \vec{b}) (\Delta \vec{b})^T & \text{ if } \kappa \Vert \Delta \vec{b} \Vert_2^2 > 1 \\ 
\vec{0} & \text{ if } \kappa \Vert \Delta \vec{b} \Vert_2^2 \leq 1
\end{cases} \ } \ .
\end{equation}
Note that this result mirrors the result we obtained in the one-dimensional case (Eq. \ref{eq:beta_1D_opt}).

When the variance-matching condition holds, the determinant of $\tilde{\vec{B}}^{-1}$ equals
\begin{equation} 
\begin{split}
\det\tilde{\vec{B}}^{-1} &= \det\left( \frac{1}{\kappa} \vec{I} + \left( 1 - \frac{1}{\kappa \Vert \Delta \vec{b} \Vert_2^2}\right) (\Delta \vec{b}) (\Delta \vec{b})^T \right) \\
 &= \det\left( \frac{1}{\kappa} \vec{I} \right) \left[ 1 + \kappa \left( 1 - \frac{1}{\kappa \Vert \Delta \vec{b} \Vert_2^2}\right)  \Vert \Delta \vec{b} \Vert_2^2 \right]\\
  &= \frac{1}{\kappa^D} \kappa \Vert \Delta \vec{b} \Vert_2^2 \\
  &= \frac{\Vert \Delta \vec{b} \Vert_2^2}{\kappa^{D-1}} \ .
\end{split}
\end{equation}
This implies that the hitting probability equals
\begin{equation}
\boxed{ \ p_{\text{hit}}(\vec{b}) := \sqrt{\frac{\det\tilde{\vec{B}}}{(2 \pi)^D }} e^{- \frac{1}{2} (  \vec{b} - \vec{b}^f  )^T \tilde{\vec{B}} (  \vec{b} - \vec{b}^f  )} = \frac{1}{\kappa^{1/2}} \left( \frac{\kappa}{2 \pi} \right)^{D/2} e^{-1/2} \ \frac{1}{\Vert \Delta \vec{b} \Vert_2} \ } \ .
\end{equation}
Our one-dimensional result (Eq. \ref{eq:hit_1D}) is a special case of this one.

\subsection{Convergence time using optimal exploration matrix}

Suppose we use the optimal curriculum given by Eq. \ref{eq:B_opt_Ddim}. How much better is the convergence time? Bias learning dynamics (Eq. \ref{eq:b_curr}) become
\begin{equation}  
\begin{split}
\frac{d}{d\tau} \vec{b} &= - \eta  g \tilde{\vec{B}} ( \Delta \vec{b}) \ \sqrt{\frac{\det\tilde{\vec{B}}}{(2 \pi)^D}} e^{- \frac{1}{2}  ( \Delta \vec{b})^T \tilde{\vec{B}}  ( \Delta \vec{b})} \\
&= - \frac{\eta g e^{-1/2}}{\kappa^{1/2}} \left( \frac{\kappa}{2 \pi} \right)^{D/2} \ \tilde{\vec{B}} \frac{\Delta \vec{b}}{\Vert \Delta \vec{b} \Vert_2} \\
&= - \frac{\eta g e^{-1/2}}{\kappa^{1/2}} \left( \frac{\kappa}{2 \pi} \right)^{D/2} \  \frac{\Delta \vec{b}}{\Vert \Delta \vec{b} \Vert_2^3} \ .
\end{split}
\end{equation}
Now define the bias error as $E := \kappa \Vert \Delta \vec{b} \Vert_2^2$. Its time derivative is
\begin{equation}
\begin{split}
\frac{d}{d\tau} E &= 2 \kappa (\Delta \vec{b})^T \frac{d}{d\tau} ( \Delta \vec{b}) \\
&= -2 \kappa \frac{\eta g e^{-1/2}}{\kappa^{1/2}} \left( \frac{\kappa}{2 \pi} \right)^{D/2} \  \frac{1}{\Vert \Delta \vec{b} \Vert_2} \\
&= -2 \kappa \eta g e^{-1/2} \left( \frac{\kappa}{2 \pi} \right)^{D/2} \  \frac{1}{\sqrt{E}} \ .
\end{split} 
\end{equation}   
We can solve this ODE via separation of variables. If error at time $t = 0$ is $E_0$ and the time at $t = t_*$ is $E_f$, then
\begin{equation}
\begin{split}
\int_{E_0}^{E_f} \sqrt{E} \ dE &= -2 \kappa \eta g e^{-1/2} \left( \frac{\kappa}{2 \pi} \right)^{D/2} \ \int_0^{t_*} dt \\
\implies \ \frac{2}{3} \left. E^{3/2} \right|_{E_0}^{E_f}  &= -2 \kappa \eta g e^{-1/2} \left( \frac{\kappa}{2 \pi} \right)^{D/2} \ t_* \ ,
\end{split}
\end{equation}
which implies that the convergence time equals
\begin{equation}
\boxed{ \ t_* = \frac{1}{3}  \frac{e^{1/2}}{\kappa \eta g} \left( \frac{2 \pi}{\kappa} \right)^{D/2} \left[ \ E_0^{3/2} - E_f^{3/2} \ \right] = \frac{1}{3}  \frac{e^{1/2}}{\kappa \eta g} \left( \frac{2 \pi}{\kappa} \right)^{D/2} \left[ \ \kappa^{3/2} \Vert \Delta \vec{b}^0 \Vert_2^3 - 1 \ \right] \ } \ ,
\end{equation}
where we chose $E_f = 1$ since this is the minimum possible value of the error where the variance-matching condition still holds.

If a smaller error is desired, we must consider a second phase of dynamics where $\tilde{\vec{B}} = \kappa \vec{I}$. Reusing the result from Appendix \ref{app:conv}, this equals
\begin{equation} 
t_*^{\text{second}} =  \frac{1}{2 \eta g \kappa} \left( \frac{2 \pi}{\kappa} \right)^{D/2} \left[  \ \text{Ei}(1/2) - \text{Ei}\left(  \frac{1}{2} \epsilon^2   \right) \ \right]   \ ,
\end{equation}
where we assume the final error equals $\epsilon^2$. 

Including both phases, the overall convergence time equals
\begin{equation}
\boxed{ \ t_* = \frac{1}{3}  \frac{e^{1/2}}{\kappa \eta g} \left( \frac{2 \pi}{\kappa} \right)^{D/2} \left[ \ \kappa^{3/2} \Vert \Delta \vec{b}^0 \Vert_2^3 - 1 \ \right] +  \frac{1}{2 \eta g \kappa} \left( \frac{2 \pi}{\kappa} \right)^{D/2} \left[  \ \text{Ei}(1/2) - \text{Ei}\left(  \frac{1}{2} \epsilon^2   \right) \ \right]  \ } \ .
\end{equation}
Importantly, this depends on the \textit{cube} of the norm of the initial bias error $\Delta \vec{b}^0$, whereas our earlier result (see Appendix \ref{app:conv}) depended \textit{exponentially} on it.

Suppose we assume, as we did in Appendix \ref{app:conv}, that the initial error scales like $E_0 = \kappa D d_0^2$. Then we have
\begin{equation}
\boxed{ \ t_* = \frac{1}{3}  \frac{e^{1/2}}{\kappa \eta g} \left( \frac{2 \pi}{\kappa} \right)^{D/2} \left[ \ \kappa^{3/2} D^{3/2} d_0^3 - 1 \ \right] +  \frac{1}{2 \eta g \kappa} \left( \frac{2 \pi}{\kappa} \right)^{D/2} \left[  \ \text{Ei}(1/2) - \text{Ei}\left(  \frac{1}{2} \epsilon^2   \right) \ \right]  \ } \ .
\end{equation}
This also scales much better with $D$ than before, since the initialization-related term goes like $D^{3/2}$. Although it still technically scales exponentially with $D$, this is essentially due to the exponential-in-$D$ scaling of the prefactor that appears in the return (see Sec. \ref{sec:form}), so this prefactor is to some extent unavoidable given our formulation.


\clearpage

\clearpage

\section{Analysis of the implicit curriculum effect}
\label{app:implicit}

In this appendix, we study the \textit{implicit curriculum effect} discussed in Sec. \ref{sec:implicit}: model weights transiently move away from their target in order to increase the hitting probability, and hence make learning faster overall. We make the same simplifying assumptions we make there:
(i) the feature covariance matrix is isotropic, i.e., $\vec{\Sigma} = s \vec{I}$ for some $s > 0$; (ii) $s > 0$ is so large that the bias parameters are effectively frozen on the time scale of weight dynamics; and (iii) the effort coefficient $\alpha/T$ is small enough that it does not appreciably influence endpoint dynamics. Given these assumptions, the endpoint weight discrepancy $\Delta \vec{W} := \vec{W} - \vec{W}^f$ approximately evolves according to
\begin{equation} \label{eq:weight_simple_app}
\boxed{ \ \frac{d}{d\tau} \Delta \vec{W} \approx - \eta s g p_{\text{hit}}  \left[    \vec{I} - \vec{V}^{-1} (\Delta \vec{b}) (\Delta \vec{b})^T  \right] \vec{V}^{-1} (\Delta \vec{W})  \ } 
\end{equation}
where, as we derived in Appendix \ref{app:dyn_deriv}, the hitting probability factor equals
\begin{equation}
p_{\text{hit}}(\vec{b}, \vec{W}) := \frac{e^{- \frac{1}{2} (\Delta \vec{b})^T \vec{V}^{-1} (\Delta \vec{b})} }{\sqrt{(2 \pi)^D \det \vec{V}}}    
\end{equation}
and the covariance matrix $\vec{V}$ equals
\begin{equation} \label{eq:V_app}
\vec{V} := \tilde{\vec{B}}^{-1}  +  (\Delta \vec{W}) \vec{\Sigma} (\Delta \vec{W})^T = \frac{1}{\tilde{\beta}} \vec{I}  +  s (\Delta \vec{W})  (\Delta \vec{W})^T  \ .
\end{equation}

\subsection{Dynamics of the effective covariance}

A useful way to analyze the implicit curriculum effect is to directly track the time evolution of $\vec{V}$ (Eq. \ref{eq:V_app}). It is straightforward to compute that
\begin{align}
\frac{d}{d\tau} \vec{V} &= - \eta s g p_{\text{hit}} \left\{ \left[    \vec{I} - \vec{V}^{-1} (\Delta \vec{b}) (\Delta \vec{b})^T  \right] \vec{V}^{-1} (\vec{V} - \tilde{\vec{B}}^{-1}) +   (\vec{V} - \tilde{\vec{B}}^{-1})   \vec{V}^{-1} \left[    \vec{I} -  (\Delta \vec{b}) (\Delta \vec{b})^T \vec{V}^{-1}  \right]     \right\} \\
&= - \eta s g p_{\text{hit}} \left\{ \left[    \vec{I} - \vec{V}^{-1} (\Delta \vec{b}) (\Delta \vec{b})^T  \right] (\vec{I} - \vec{V}^{-1}  \tilde{\vec{B}}^{-1}) +   (\vec{I} - \tilde{\vec{B}}^{-1}  \vec{V}^{-1})   \left[    \vec{I} -  (\Delta \vec{b}) (\Delta \vec{b})^T \vec{V}^{-1}  \right]     \right\} \notag \ .
\end{align}
Since we are assuming that $\tilde{\vec{B}} = \tilde{\beta} \vec{I}$, we can specialize this to
\begin{align} \label{eq:V_dyn_app}
\frac{d}{d\tau} \vec{V} &= - \eta s g p_{\text{hit}} \left\{ \left[    \vec{I} - \vec{V}^{-1} (\Delta \vec{b}) (\Delta \vec{b})^T  \right] (\vec{I} - \frac{1}{\tilde{\beta}}\vec{V}^{-1} ) +   (\vec{I} - \frac{1}{\tilde{\beta}} \vec{V}^{-1})   \left[    \vec{I} -  (\Delta \vec{b}) (\Delta \vec{b})^T \vec{V}^{-1}  \right]     \right\} \\
&= - 2 \eta s g p_{\text{hit}} \left\{  \vec{I} - \frac{1}{\tilde{\beta}} \vec{V}^{-1} + \frac{1}{\tilde{\beta}}  \vec{V}^{-1} (\Delta \vec{b}) (\Delta \vec{b})^T \vec{V}^{-1}   - \frac{1}{2} \vec{V}^{-1} (\Delta \vec{b}) (\Delta \vec{b})^T - \frac{1}{2}  (\Delta \vec{b}) (\Delta \vec{b})^T \vec{V}^{-1}  \right\} \notag \ .
\end{align}

\subsection{Fixed point of covariance dynamics}

The covariance matrix $\vec{V}$ stops changing when the right-hand side of Eq. \ref{eq:V_dyn_app} equals zero, i.e., when
\begin{equation}
 \vec{I} - \frac{1}{\tilde{\beta}} \vec{V}^{-1} + \frac{1}{\tilde{\beta}}  \vec{V}^{-1} (\Delta \vec{b}) (\Delta \vec{b})^T \vec{V}^{-1}   = \frac{1}{2} \vec{V}^{-1} (\Delta \vec{b}) (\Delta \vec{b})^T +  \frac{1}{2} (\Delta \vec{b}) (\Delta \vec{b})^T \vec{V}^{-1} \ .
\end{equation}
This is an algebraic Riccati equation. To ease notation, use $\vec{M}$ to denote $\vec{V}^{-1}$. Let $\vec{e}$ denote the unit vector parallel to $\Delta \vec{b}$ and $r$ the magnitude of $\Delta \vec{b}$, so that $\Delta \vec{b} = r \vec{e}$. Finally, let $\vec{u}$ denote the unit vector parallel to $\vec{V}^{-1} \vec{e} = \vec{M} \vec{e}$ and $v$ the magnitude of $\vec{M} \vec{e}$. 

Given this change in notation, our algebraic Riccati equation becomes
\begin{equation}
\boxed{ \ \vec{I} - \frac{1}{\tilde{\beta}} \vec{M} + \frac{r^2 v^2}{\tilde{\beta}} \vec{u} \vec{u}^T  = \frac{r^2 v}{2} \vec{u} \vec{e}^T + \frac{r^2 v}{2} \vec{e} \vec{u}^T  \ } \ .
\end{equation}
\paragraph{Simplifying Riccati equation.} Left- and right-multiplying by $\vec{e}$ yields
\begin{equation} \label{eq:re_riccati}
1 - \frac{1}{\tilde{\beta}} v (\vec{u}^T \vec{e}) + \frac{r^2 v^2}{\tilde{\beta}}  (\vec{u}^T \vec{e})^2  = r^2 v (\vec{u}^T \vec{e})  \ . 
\end{equation}
Right-multiplying by $\vec{e}$ and left-multiplying by $\vec{u}$ yields
\begin{equation}
 \vec{u}^T\vec{e} - \frac{1}{\tilde{\beta}} v  + \frac{r^2 v^2}{\tilde{\beta}}  (\vec{u}^T \vec{e})  = \frac{r^2 v}{2}   + \frac{r^2 v}{2}  (\vec{u}^T \vec{e})^2 \ .
\end{equation}
Rearranging, we have the two conditions
\begin{equation}
\begin{split}
0 &= \frac{r^2 v^2}{\tilde{\beta}}  (\vec{u}^T \vec{e})^2 - \left[ \frac{1}{\tilde{\beta}} v + r^2 v \right] (\vec{u}^T \vec{e}) + 1  \\
0 &=  \frac{r^2 v}{2}  (\vec{u}^T \vec{e})^2 - \left[ 1 + \frac{r^2 v^2}{\tilde{\beta}} \right] (\vec{u}^T \vec{e}) + \left[ \frac{r^2 v}{2} + \frac{v}{\tilde{\beta}} \right]
\end{split}
\end{equation}

\paragraph{The bias error is an eigenvector at the fixed point.} We will first use these conditions to show that $\vec{u}^T \vec{e} = \pm 1$, i.e., that $\vec{u}$ and $\vec{e}$ are parallel, which implies that $\vec{e}$ is an eigenvector of $\vec{M}$. Using the quadratic formula to solve the first equation, we find that
\begin{equation}
\begin{split}
\vec{u}^T \vec{e} &= \frac{ \tilde{\beta}}{2 r^2 v^2 } \left\{ \left[ \frac{1}{\tilde{\beta}} v + r^2 v \right] \pm \sqrt{ \left[ \frac{1}{\tilde{\beta}} v + r^2 v \right]^2 - \frac{4 r^2 v^2}{\tilde{\beta}}} \right\} \\
&= \frac{\tilde{\beta}}{2 r^2 v^2 } \left\{ \left[ \frac{1}{\tilde{\beta}} v + r^2 v \right] \pm \left[ \frac{1}{\tilde{\beta}} v - r^2 v \right] \right\} \ ,
\end{split}
\end{equation}
i.e., that the two solutions are
\begin{equation}
\vec{u}^T \vec{e} = \frac{1}{r^2 v} \ , \frac{\tilde{\beta}}{v} \ .
\end{equation}
If $\vec{u}^T \vec{e} = 1/(r^2 v)$, then the second condition becomes
\begin{equation}
0 =  \frac{1}{2 r^2 v}   - \left[ 1 + \frac{r^2 v^2}{\tilde{\beta}} \right] \frac{1}{r^2 v} + \left[ \frac{r^2 v}{2} + \frac{v}{\tilde{\beta}} \right] = - \frac{1}{2 r^2 v} + \frac{r^2 v}{2} \ ,
\end{equation}
which implies $r^2 v = 1/(r^2 v)$, i.e., that $r^2 v = 1$. Equivalently, $\vec{u}^T \vec{e} = \pm 1$.

If we instead take the other solution $\vec{u}^T \vec{e} = \tilde{\beta}/v$ to be true, the second condition becomes 
\begin{equation}
\begin{split}
0 &=  \frac{r^2 v}{2} \frac{\tilde{\beta}^2}{v^2} - \left[ 1 + \frac{r^2 v^2}{\tilde{\beta}} \right]\frac{\tilde{\beta}}{v} + \left[ \frac{r^2 v}{2} + \frac{v}{\tilde{\beta}} \right] \\
&= \frac{r^2 }{2} \frac{\tilde{\beta}^2}{v} - \frac{\tilde{\beta}}{v} +  \frac{v}{\tilde{\beta}}  - \frac{1}{2} r^2 v  \\
&= \frac{\tilde{\beta}}{v} \left( \frac{r^2 }{2} \tilde{\beta} - 1 \right) - \frac{v}{\tilde{\beta}} \left( \frac{r^2 }{2} \tilde{\beta} - 1 \right) \\
&= \left( \frac{r^2 }{2} \tilde{\beta} - 1 \right) \left(    \frac{\tilde{\beta}}{v}  - \frac{v}{\tilde{\beta}} \right) \ .
\end{split}
\end{equation}
Importantly, we would like equality to hold \textit{regardless} of $r$. Since in most cases it will not be true that $r^2 = 2/\tilde{\beta}$, we ought to take $v^2 = \tilde{\beta}^2$. But then $\vec{u}^T \vec{e} = \pm 1$. Hence, either way we obtain that $\vec{u}$ and $\vec{e}$ are parallel, so $\vec{e}$ is an eigenvector of $\vec{M}$ when the algebraic Riccati equation above holds.

\paragraph{Expression for the fixed point.} If $\vec{e}$ is an eigenvector of $\vec{M}$, the Riccati equation above (Eq. \ref{eq:re_riccati} reduces to
\begin{equation}
 \vec{I} - \frac{1}{\tilde{\beta}} \vec{M} + \frac{r^2 v^2}{\tilde{\beta}} \vec{e} \vec{e}^T  = r^2 v \ \vec{e} \vec{e}^T \ ,
\end{equation}
which implies
\begin{equation}
\vec{M} = \tilde{\beta} \vec{I} + r^2 v \left[ v - \tilde{\beta} \right] \vec{e} \vec{e}^T \ .
\end{equation}
Note that $v$ is the eigenvalue associated with $\vec{e}$, since (given how we have defined things) $\vec{M} \vec{e} = v \vec{u} = v \vec{e}$. Consistency requires that
\begin{equation}
v = \tilde{\beta} + r^2 v (v - \tilde{\beta}) \ ,
\end{equation}
which is equivalent to the quadratic equation
\begin{equation}
r^2 v^2 - \left[ 1 + \tilde{\beta} r^2 \right] v + \tilde{\beta} = 0 \ .
\end{equation}
The solutions of this equation are
\begin{equation}
v = \frac{1}{2 r^2} \left\{ 1 + \tilde{\beta} r^2  \pm \sqrt{ \left[ 1 + \tilde{\beta} r^2 \right]^2 - 4 r^2 \tilde{\beta} }  \right\} = \frac{1}{2 r^2} \left\{ 1 + \tilde{\beta} r^2  \pm  \left[ 1 -\tilde{\beta} r^2 \right]  \right\}  \ ,
\end{equation}
or
\begin{equation}
v = \frac{1}{r^2} \ , \tilde{\beta} \ .
\end{equation}
These two choices of $v$ correspond to qualitatively different solutions. If $v = \tilde{\beta}$, then $\vec{M} = \tilde{\beta} \vec{I}$ is proportional to the identity matrix. But the other $v$ solution implies
\begin{equation}
\vec{M} = \tilde{\beta} \vec{I} +  \left[ \frac{1}{r^2} - \tilde{\beta} \right] \vec{e} \vec{e}^T \ .
\end{equation}
This solution is not valid for \textit{all} possible values of $r$. To see this, note that since $\vec{M} = \vec{V}^{-1}$, we have
\begin{equation}
\vec{V} = \frac{1}{\tilde{\beta}} \left[  \vec{I} + \left[ \frac{1}{\tilde{\beta} r^2} - 1 \right] \vec{e} \vec{e}^T \right]^{-1} = \frac{1}{\tilde{\beta}} \left[  \vec{I} - \tilde{\beta} r^2 \left[ \frac{1}{\tilde{\beta} r^2} - 1 \right] \vec{e} \vec{e}^T \right] =  \frac{1}{\tilde{\beta}} \left[  \vec{I} -  \left[ 1 - \tilde{\beta} r^2 \right] \vec{e} \vec{e}^T \right] \ .
\end{equation}
Now since $\vec{V} = (1/\tilde{\beta}) \vec{I} + (\Delta \vec{W}) (\Delta \vec{W})^T$, it must be true that
\begin{equation}
(\Delta \vec{W}) (\Delta \vec{W})^T = - \frac{1}{\tilde{\beta}} \left[ 1 - \tilde{\beta} r^2 \right] \vec{e} \vec{e}^T 
\end{equation}
is positive semidefinite. But this can only happen when
\begin{equation}
\tilde{\beta} r^2 \geq 1 \ .
\end{equation}
If we change notation, this is just
\begin{equation}
\tilde{\beta} \Vert \Delta \vec{b} \Vert_2^2 \geq 1 \ ,
\end{equation}
i.e., the condition we encountered in Sec. \ref{sec:curr} and Appendix \ref{app:optimal} (note that if $\vec{B}$ is controllable, we should replace $\tilde{\beta}$ with $\kappa$). In words: the agent's bias must be far enough from the target bias that increasing exploration can increase the hitting probability.

We conclude that there are two fixed points of $\vec{V}$. One has $\vec{V} = (1/\tilde{\beta}) \vec{I}$ (i.e., the weight discrepancy is zero), and the other has
\begin{equation} \label{eq:V_nontrivial}
\vec{V} = \frac{1}{\tilde{\beta}} \vec{I} + \left( 1 - \frac{1}{\tilde{\beta} \Vert \Delta \vec{b} \Vert_2^2}  \right) (\Delta \vec{b}) (\Delta \vec{b})^T 
\end{equation}
but requires that $\tilde{\beta} \Vert \Delta \vec{b} \Vert_2^2 \geq 1$. It is important to point out that this result is essentially the same as Eq. \ref{eq:B_opt_Ddim}, the optimal choice of $\vec{B}^{-1}$.

\subsection{Error direction gets magnified while other directions decay}

Let us assume for now that $\Delta \vec{b}$ is an eigenvector of $\vec{V}$ (and $\vec{V}^{-1}$), since, as we just found, this is true at the fixed point. Since $\vec{V}$ is positive definite, and in particular symmetric, its other eigenvectors are orthogonal to $\Delta \vec{b}$. 

Let $\lambda$ denote the eigenvalue of $\vec{V}$ corresponding to $\Delta \vec{b}$. Let $\vec{v}$ denote an eigenvector orthogonal to $\Delta \vec{b}$ and $\lambda_{\perp}$ its corresponding eigenvalue. Then our equation for $\vec{V}$ dynamics (Eq. \ref{eq:V_dyn_app}) implies
\begin{equation}
\begin{split}
\frac{d}{d\tau} \lambda &= - \eta s g p_{\text{hit}} \left\{ 2 \left[ 1 - \frac{1}{\tilde{\beta}} \frac{1}{\lambda} + \frac{1}{\tilde{\beta}} \frac{\Vert \Delta \vec{b} \Vert_2^2}{\lambda^2}  \right]  - 2 \frac{\Vert \Delta \vec{b} \Vert_2^2}{\lambda} \right\} \\
&= - 2 \eta s g p_{\text{hit}} \ \left(  1 - \frac{1}{\tilde{\beta}} \frac{1}{\lambda}  \right) \left(  1 - \frac{\Vert \Delta \vec{b} \Vert_2^2}{\lambda} \right) \ .
\end{split}
\end{equation}
This system has two fixed points:
\begin{equation}
\lambda = \frac{1}{\tilde{\beta}} \ , \ \Vert \Delta \vec{b} \Vert_2^2 \ .
\end{equation}
The one with $\lambda = 1/\tilde{\beta}$ corresponds to the trivial fixed point with $\vec{V} = (1/\tilde{\beta}) \vec{I}$. The other corresponds to the nontrivial fixed point (Eq. \ref{eq:V_nontrivial}) at which the variance-matching condition holds. 

Importantly, it is clear from the above ODE that the trivial fixed point is \textit{unstable} if it is true that $\tilde{\beta} \Vert \Delta \vec{b} \Vert_2^2 > 1$; hence, the nontrivial fixed point is the only stable one in that case.

Meanwhile, eigenvectors orthogonal to $\Delta \vec{b}$ have dynamics
\begin{equation}
\frac{d}{d\tau} \lambda_{\perp} = - 2  \eta s g p_{\text{hit}} \left( 1 - \frac{1}{\tilde{\beta} \lambda_{\perp}} \right)
\end{equation}
with only a single fixed point at $\lambda_{\perp} = 1/\tilde{\beta}$. In other words, the eigenvalues of $\vec{V}$ associated with orthogonal directions only decay toward the fixed point where the corresponding $\Delta \vec{W}$ component equals zero.

\subsection{Evolution of eigenbasis}

Suppose $\Delta \vec{b}$ is \textit{not} an eigenvector of $\vec{V}$. Can we still say something about its approach to either fixed point? It turns out that, if we explicitly compute the dynamics of the eigenbasis of $\vec{V}$, we can show that one eigenvector generally orients in the direction of $\Delta \vec{b}$.

Assume $\vec{V} = \vec{U} \vec{\Lambda} \vec{U}^T$. Then
\begin{equation}
\frac{d}{d\tau} \vec{V} = \dot{\vec{U}} \vec{\Lambda} \vec{U}^T + \vec{U} \dot{\vec{\Lambda}} \vec{U}^T + \vec{U} \vec{\Lambda} \dot{\vec{U}}^T  \ .
\end{equation}
Relevant facts are that
\begin{equation}
\vec{U} \vec{U}^T = \vec{I} \ \implies \ \dot{\vec{U}} \vec{U}^T = - \vec{U} \dot{\vec{U}}^T \ \implies \ \sum_k \dot{U}_{i k} U_{j k} = - \sum_k U_{i k} \dot{U}_{j k}
\end{equation}
\begin{equation}
\vec{U}^T \vec{U} = \vec{I} \ \implies \ \dot{\vec{U}}^T \vec{U} = - \vec{U}^T \dot{\vec{U}} \ \implies \ \sum_k \dot{U}_{k i} U_{k j} = - \sum_k U_{k i} \dot{U}_{k j} \ .
\end{equation}
Using the above results, and the additional fact that  $\vec{V}^{-1} = \vec{U} \vec{\Lambda}^{-1} \vec{U}^T$, we can rewrite $\vec{V}$ dynamics (Eq. \ref{eq:V_dyn_app}) as
\begin{equation*}
\begin{split}
\frac{d}{d\tau} \vec{V} &= - \eta s g p_{\text{hit}} \vec{U} \left\{ 2 \left[  \vec{I} - \frac{1}{\tilde{\beta}} \vec{\Lambda}^{-1} + \frac{1}{\tilde{\beta}}  \vec{\Lambda}^{-1} \vec{U}^T (\Delta \vec{b}) (\Delta \vec{b})^T \vec{U} \vec{\Lambda}^{-1}  \right]  - \vec{\Lambda}^{-1} \vec{U}^T (\Delta \vec{b}) (\Delta \vec{b})^T \vec{U} -  (\Delta \vec{b}) (\Delta \vec{b})^T \vec{U} \vec{\Lambda}^{-1}  \right\} \vec{U}^T .
\end{split}
\end{equation*}
Define $\vec{y} := \vec{U}^T \Delta \vec{b}$. Then
 \begin{equation*}
\begin{split}
\frac{d}{d\tau} \vec{V} &= - \eta s g p_{\text{hit}} \vec{U} \left\{ 2 \left[  \vec{I} - \frac{1}{\tilde{\beta}} \vec{\Lambda}^{-1} + \frac{1}{\tilde{\beta}}  \vec{\Lambda}^{-1}   \vec{y} \vec{y}^T \vec{\Lambda}^{-1}  \right]  - \vec{\Lambda}^{-1}   \vec{y} \vec{y}^T  -  \vec{y} \vec{y}^T \vec{\Lambda}^{-1}  \right\} \vec{U}^T \ .
\end{split}
\end{equation*}
Putting everything together, we have
 \begin{equation*}
\begin{split}
\vec{U}^T \dot{\vec{U}}  \vec{\Lambda}  +  \dot{\vec{\Lambda}} + \vec{\Lambda} \dot{\vec{U}}^T \vec{U}  &= - \eta s g p_{\text{hit}}  \left\{ 2 \left[  \vec{I} - \frac{1}{\tilde{\beta}} \vec{\Lambda}^{-1} + \frac{1}{\tilde{\beta}}  \vec{\Lambda}^{-1}   \vec{y} \vec{y}^T \vec{\Lambda}^{-1}  \right]  - \vec{\Lambda}^{-1}   \vec{y} \vec{y}^T  -  \vec{y} \vec{y}^T \vec{\Lambda}^{-1}  \right\} \ .
\end{split}
\end{equation*}
\paragraph{Eigenvalue dynamics.} The $(i, j)$ component of the left-hand side equals
\begin{equation}
\sum_{k} U_{k i} \dot{U}_{k j} \lambda_j + \delta_{i j} \dot{\lambda}_i + \sum_{k} \dot{U}_{k i} U_{k j} \lambda_i  \ ,
\end{equation}
so its diagonal equals $\delta_{i j} \dot{\lambda}_i $because the other parts cancel. This means
\begin{equation} \label{eq:eig_gen}
\dot{\lambda}_i = - 2 \eta s g p_{\text{hit}}  \left\{  1 - \frac{1}{\tilde{\beta}} \frac{1}{\lambda_i} + \frac{1}{\tilde{\beta}} \frac{y_i^2}{\lambda_i^2}  -  \frac{y_i^2}{\lambda_i} \right\} = - 2 \eta s g p_{\text{hit}}  \left(  1 - \frac{y_i^2}{\lambda_i} \right) \left( 1 - \frac{1}{\tilde{\beta} \lambda_i}   \right)\ .
\end{equation}
Note that this is almost what we had before for the eigenvector aligned with $\Delta \vec{b}$, except this time the component $y_i$ of $\Delta \vec{b}$ along this (not necessarily aligned) eigenvector appears instead.

\paragraph{Eigenvector dynamics.} Meanwhile, the off-diagonal entries are
\begin{equation}
\begin{split}
\sum_{k} U_{k i} \dot{U}_{k j} \lambda_j + \sum_{k} \dot{U}_{k i} U_{k j} \lambda_i &= - 2 \eta s g p_{\text{hit}}  \left\{  \frac{1}{\tilde{\beta}} \frac{y_i y_j}{\lambda_i \lambda_j}  -  \frac{1}{2} \left( \frac{1}{\lambda_i} + \frac{1}{\lambda_j} \right) y_i y_j \right\} \\
&= -  \eta s g p_{\text{hit}}  \left\{  \frac{2}{\tilde{\beta}}  -   \left(\lambda_i + \lambda_j \right) \right\} \frac{y_i y_j}{\lambda_i \lambda_j} \ .
\end{split}
\end{equation}
Define $\vec{\Omega} := \vec{U}^T \dot{\vec{U}}$. Then
\begin{equation}
\vec{\Omega}^T = \dot{\vec{U}}^T \vec{U} = - \vec{U}^T \dot{\vec{U}} = - \vec{\Omega}  \ ,
\end{equation}
so we can rewrite the off-diagonal dynamics as
\begin{equation}
\Omega_{i j} (\lambda_j - \lambda_i) = -  \eta s g p_{\text{hit}}  \left\{  \frac{2}{\tilde{\beta}}  -   \left(\lambda_i + \lambda_j \right) \right\} \frac{y_i y_j}{\lambda_i \lambda_j} \ .
\end{equation}
Now dynamics of $\vec{y}$ follow
\begin{equation}
\frac{d}{d\tau} \vec{y} = \frac{d}{d\tau} \vec{U}^T \Delta \vec{b} = \dot{\vec{U}}^T \vec{U} \vec{y} = - \vec{\Omega} \vec{y} \ ,
\end{equation}
or in terms of components,
\begin{equation} \label{eq:y_dyn}
\dot{y}_i = - \sum_{j \neq i}  \eta s g p_{\text{hit}}  \left\{  \frac{2}{\tilde{\beta}}  -   \left(\lambda_i + \lambda_j \right) \right\} \frac{1}{(\lambda_i - \lambda_j)} \frac{y_i y_j^2}{\lambda_i \lambda_j} \ .
\end{equation}
\paragraph{Winner-take-all dynamics.} Because $\vec{\Omega}$ is skew-symmetric, there is a conservation law: the norm of $\vec{y}$ never changes. To see this, observe that
\begin{equation}
\frac{d}{d\tau} \frac{\vec{y}^T \vec{y}}{2} = - \vec{y}^T \vec{\Omega}^T \vec{y} - \vec{y}^T \vec{\Omega} \vec{y} = 0 \ .
\end{equation}
Hence, the sum of the squares of the $y_i$ must always add up to
\begin{equation}
\vec{y}^T \vec{y}  = (\Delta \vec{b})^T \vec{U} \vec{U}^T (\Delta \vec{b}) =  \Vert \Delta \vec{b} \Vert_2^2 \ .
\end{equation}
This has the following consequence. If one of the $y_i$ is somewhat larger than the rest, the dynamics of its corresponding eigenvalue $\lambda_i$ (Eq. \ref{eq:eig_gen}) will tend to ensure that $\lambda_i \approx y_i^2$. If the other eigenvalues are somewhat small, for example because the corresponding $y_j$ are small, then the dynamics of the larger $y_i$ (Eq. \ref{eq:y_dyn}) will likely have a positive right-hand side. Hence, the larger $y_i$ will get even larger, until eventually $|y_i| \approx \Vert \Delta \vec{b} \Vert_2$. In this sense, we obtain `winner-take-all' dynamics when there is an imbalance in the $y_i$. 

We can see this most explicitly in the 2D case. If we only have $y_i$ and $y_j$ with $y_i^2 + y_j^2 =  \Vert \vec{y} \Vert_2^2 $, then there are two ODEs
\begin{equation}
\begin{split}
\dot{y}_i &= -   \eta s g p_{\text{hit}}  \left\{  \frac{2}{\tilde{\beta}}  -   \left(\lambda_i + \lambda_j \right) \right\} \frac{1}{(\lambda_i - \lambda_j)} \frac{y_i y_j^2}{\lambda_i \lambda_j} \\
\dot{y}_j &= -   \eta s g p_{\text{hit}}  \left\{  \frac{2}{\tilde{\beta}}  -   \left(\lambda_i + \lambda_j \right) \right\} \frac{1}{(\lambda_j - \lambda_i)} \frac{y_j y_i^2}{\lambda_i \lambda_j} \ .
\end{split}
\end{equation}
By our conservation law, $y_j^2 = \Vert \vec{y} \Vert_2^2 - y_i^2$, so
\begin{equation} \label{eq:winner_2}
\begin{split}
\dot{y}_i &=    \eta s g p_{\text{hit}}  \left\{      \lambda_i + \lambda_j - \frac{2}{\tilde{\beta}}   \right\} \frac{1}{(\lambda_i - \lambda_j)} \frac{y_i (y^2 - y_i^2)}{\lambda_i \lambda_j} \ .
\end{split}
\end{equation}
If $\Vert \vec{y} \Vert_2^2$ is bigger than $2/\tilde{\beta}$, and $y_i^2 \neq y_j^2$, then at least one $y_i^2$ is bigger than $1/\tilde{\beta}$ . Then at least one $\lambda_i$ will be pushed toward the nontrivial fixed point. Then Eq. \ref{eq:winner_2} indicates that the bigger $y_i$ will increase until $y_i^2 = \Vert \vec{y} \Vert_2^2$. 


\end{document}